\documentclass[11pt]{article}

\usepackage[final]{acl}

\usepackage{times}
\usepackage{latexsym}

\usepackage[T1]{fontenc}

\usepackage[utf8]{inputenc}

\usepackage{microtype}

\usepackage{inconsolata}

\usepackage{graphicx}

\usepackage[ruled,vlined]{algorithm2e}

\usepackage{booktabs}
\usepackage{siunitx}
\usepackage[table]{xcolor}
\usepackage{multirow}
\usepackage{subcaption}
\usepackage{amssymb}
\usepackage{amsmath}
\usepackage{tabularx}
\usepackage{booktabs}
\usepackage{listings}
\newcommand{\ms}[2]{\ensuremath{#1_{\pm #2}}}
\usepackage{tikz}
\usetikzlibrary{arrows.meta, decorations.pathreplacing, positioning}

\definecolor{topval}{RGB}{198,239,206}

\usepackage{caption}    
\usepackage{wrapfig}
\usepackage{adjustbox}
\usepackage{float}

\definecolor{color_1}{HTML}{3B82F6} 
\definecolor{color_2}{HTML}{F59E0B} 
\definecolor{color_3}{HTML}{10B981} 
\definecolor{color_4}{HTML}{8B5CF6} 
\definecolor{color_5}{HTML}{6B7280} 

\usepackage{hyperref}
\usepackage{url}

\title{PRISM: A Dual View of LLM Reasoning through Semantic Flow and Latent Computation}

\author{
  Ruidi Chang$^{1}$ \quad
  Jiawei Zhou$^{2}$ \quad
  Hanjie Chen$^{1}$ \\
  $^{1}$Rice University \quad
  $^{2}$Stony Brook University \\
  ruidi.chang@rice.edu, jiawei.zhou.1@stonybrook.edu, hanjie@rice.edu \\
}

\begin{document}
\maketitle
\begin{abstract}
Large language models (LLMs) solve complex problems by generating multi-step reasoning traces. Yet these traces are typically analyzed from only one of two perspectives: the sequence of tokens across different reasoning steps in the generated text, or the hidden-state vectors across model layers within one step. We introduce \textit{PRISM} (\textbf{P}robabilistic \textbf{R}easoning \textbf{I}nspection through \textbf{S}emantic and \textbf{I}mplicit \textbf{M}odeling), a framework and diagnostic tool for jointly analyzing both levels, providing a unified view of how reasoning evolves across steps and layers. 
Across multiple reasoning models and benchmarks, PRISM uncovers systematic patterns in the reasoning process, showing that failed trajectories are more likely to become trapped in unproductive verification loops and further diverge into distinct modes such as \textit{overthinking} and \textit{premature commitment}, which behave differently once a candidate answer is reached. It further reveals how prompting reshapes reasoning behavior beyond aggregate accuracy by altering both semantic transitions and internal computational patterns. By modeling reasoning trajectories as structured processes, PRISM makes these behaviors observable and analyzable rather than relying solely on final-task accuracy.
Taken together, these insights position PRISM as a practical tool for analyzing and diagnosing reasoning processes in LLMs. Demo and code are available at \href{https://github.com/chili-lab/PRISM}{https://github.com/chili-lab/PRISM}.
\end{abstract}

\section{Introduction}

\begin{figure*}
    \centering
    \includegraphics[width=\linewidth]{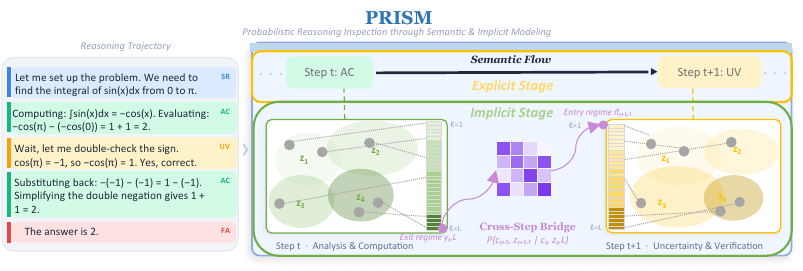}
    \caption{PRISM: an \textit{explicit stage} modeling semantic category flows (Section~\ref{sec:explicit}), where each reasoning step is assigned a semantic role such as Setup \& Retrieval (SR), Analysis \& Computation (AC), Uncertainty \& Verification (UV), or Final Answer (FA); an \textit{implicit stage} capturing computational structures within each step via Gaussian mixture models, producing latent computational regimes $z_1, z_2, \ldots$, with a cross-step bridge matrix that jointly models category transitions and internal computational patterns between consecutive steps.}
    \label{fig:framework}
\end{figure*}

Large language models (LLMs) have demonstrated strong capabilities in mathematics, scientific reasoning, and deliberative problem solving \citep{jaech2024openai, chen2025reasoning}.
However, aggregate accuracy indicates only whether the final answer is correct, offering little insight into how the model arrives there. In this paper, we study reasoning trajectories: 
reasoning in language models unfolds through a two-dimensional process. At each generation step, the model computes a sequence of hidden-state vectors across layers, transforming the representation from layer to layer along the model depth dimension and producing a token prediction at the final layer. Along the time dimension with multiple steps, these predicted tokens form a sequence in the generated text, which constitutes a visible reasoning trajectory that carry the semantic information in natural language for problem solving, leading to the final answer.
This gives rise to two complementary views of reasoning: \textit{implicit reasoning}, reflected in hidden-state vectors across layers within each step, and \textit{explicit reasoning}, reflected in semantic flow in token sequences across steps in the generated text~\cite{li2025implicit}.
Yet one fundamental question remain unresolved: \textit{how can we comprehensively characterize reasoning across both its internal structural dynamics and its external semantic progression?}

Recent studies suggest that reasoning effectiveness depends not only on the correctness of individual steps but also on their structural roles, with a few ``anchor'' steps disproportionately shaping outcomes \citep{bogdan2025thought}. Mechanistic interpretability work demonstrates that reasoning in LLMs does not emerge homogeneously across all layers or components, but is carried out by specific architectural modules \citep{cabannes2024iteration, dutta2024think}. However, these two perspectives are rarely integrated into a unified analytical framework for understanding reasoning trajectories.

To address this, we introduce \textit{PRISM} (\textbf{P}robabilistic \textbf{R}easoning \textbf{I}nspection through \textbf{S}emantic and \textbf{I}mplicit \textbf{M}odeling), an analytical framework and diagnostic tool for analyzing reasoning trajectories across both the explicit and implicit stages (Figure~\ref{fig:framework}). PRISM models the observable flow of reasoning steps with a Markov transition process~\cite{billingsley1961statistical}, captures latent computational regimes within each step using category-specific Gaussian mixture models~\cite{bishop2006pattern}, and links consecutive steps through a bridge matrix that models how the latent computational state of the previous step influences the next step. Together, these two stages provide a unified view linking the semantic role of a reasoning step with the internal computational patterns that realize it, enabling a structured analysis of both the external trajectory and the internal computation.

Across four open-source reasoning models and four benchmarks, PRISM provides three key insights. 
First, correct and incorrect reasoning traces exhibit distinct semantic dynamics: failed trajectories enter the verification stage more readily but struggle to return to computation, making verification a sticky state that stalls progress. 
Second, failed trajectories exhibit distinct behaviors in how candidate answers are reached and handled, revealing patterns consistent with overthinking and premature commitment. 
Third, PRISM enables analysis of prompting beyond final accuracy, revealing how prompt changes reshape both semantic transitions and internal computational patterns, which in turn affect end-task accuracy.
Together, these contributions position PRISM as both an interpretability framework and a practical tool for analyzing reasoning behavior in LLMs.
\section{Methodology}
\label{methdology}

PRISM characterizes reasoning through three components: an \textit{explicit stage} modeling semantic category flows (Section~\ref{sec:explicit}), an \textit{implicit stage} capturing computational structures within each step via Gaussian mixture models, and a \textit{cross-step bridge matrix} that jointly models category transitions and internal computational patterns between consecutive steps, see Figure~\ref{fig:framework}. The implicit stage and bridge are jointly trained via a two-phase Expectation--Maximization (EM) procedure~\cite{bishop2006pattern,dempster1977maximum}.

We segment generated tokens into reasoning steps $\{s_1,\dots,s_T\}$ using blank-line delimiters and extract the first-token hidden representation of each step at every transformer layer, yielding $H \in \mathbb{R}^{L\times T \times d}$, where $L$ is the number of layers, $T$ is the number of steps and $d$ is the hidden dimension. We apply normalization across layers and steps, followed by a PCA projection, to compress each hidden representation $h_{t,\ell}\in \mathbb{R}^{d}$ to $d_{\mathrm{pca}}$ dimensions ($d_{\mathrm{pca}} \ll d$) for stable estimation (Appendix~\ref{app:method}). All subsequent analyses operate on the resulting $\hat{h}_{t,\ell} \in \mathbb{R}^{d_{\mathrm{pca}}}$, where $t \in \{1,\dots,T\}$ is the step index and $\ell \in \{1,\dots,L\}$ is the layer index.

\subsection{Explicit Reasoning Transitions}
\label{sec:explicit}

We model reasoning as a sequence of semantic transitions through four categories: \texttt{final\_answer} (FA), \texttt{setup\_and\_retrieval} (SR), \texttt{analysis\_and\_computation} (AC), and \texttt{uncertainty\_and\_verification} (UV), plus an \texttt{unknown} fallback. Each step $s_t$ is classified by Qwen2.5-32B-Instruct~\cite{qwen2,qwen2.5} (see Appendix~\ref{app:method}) into $c_t \in \mathcal{C}$, $|\mathcal{C}|=4$.

The category sequence $c_{1:T}$ is modeled as a $m$-th order Markov chain. For order $m$, the transition probability conditioned on the preceding $m$ categories is
\begin{equation}
p(c_t \mid c_{t-m:t-1}) = A^{(m)}_{(c_{t-m},\dots,c_{t-1}),\, c_t},
\end{equation}
where $A^{(m)} \in \mathbb{R}^{|\mathcal{C}|^m \times |\mathcal{C}|}$ is a row-stochastic matrix whose rows index context $m$-tuples, estimated by maximum-likelihood from empirical transitions:
\begin{equation}
\begin{split}
A^{(m)}_{u,\,c}
  &= \frac{\#\bigl\{t : (c_{t-m},\dots,c_{t-1})=u,\; c_t=c\bigr\}}
          {\#\bigl\{t : (c_{t-m},\dots,c_{t-1})=u\bigr\}},
\end{split}
\end{equation}
where $u = (c_{t-m},\dots,c_{t-1}) \in \mathcal{C}^m$ denotes a length-$m$ context tuple.
The first-order case ($m{=}1$) yields the standard $|\mathcal{C}| \times |\mathcal{C}|$ transition matrix; higher orders capture context-dependent patterns.

\subsection{Implicit Reasoning Regimes and Cross-Step Bridges}
\label{sec:implicit}

\paragraph{Overview.}
To characterize the internal computation patterns associated with each reasoning category, the implicit stage models the layer-wise hidden states of each reasoning step. The input to this stage is the set of PCA-projected hidden states $\{\hat{h}_{t,\ell}\}_{\ell=1}^{L}$. The goal is to fit these layer-wise representations with a small number of latent computation modes. To do this, for each category we fit a category-specific Gaussian mixture model (GMM)~\cite{bishop2006pattern}. The GMM separates the geometric space into different clusters based on the hidden vectors. Each cluster corresponds to a latent \emph{reasoning regime}, and each layer hidden state is characterized by a probability distribution over these regimes. The output of this stage is therefore a structured description of how a reasoning step is internally realized across Transformer depth: for each layer, we obtain its posterior distribution over latent regimes.
To further connect consecutive reasoning steps, we introduce a \emph{bridge matrix}, which models how the regime at the end of one step influences the regime at the beginning of the next jointly with the category transition.

\paragraph{Model.}
Our training objective is to find a set of regimes whose mixture best fits the observed layer-wise hidden states. Inspired by GMMs, we formulate this as a joint likelihood $p(\hat{h}_{t,1:L} \mid c_t{=}c,\, \gamma_{t-1,L})$ of all layers of step $t$ with category $c_t{=}c$, conditioned on the exit posterior $\gamma_{t-1,L}$ of the preceding step. The joint likelihood has four components: (1) regime emission densities $b^{(c)}_k(\hat{h}_{t,\ell})$, which give the density of hidden state $\hat{h}_{t,\ell}$ conditioned on regime $k$; (2) mixture weights $\pi^{(c)} \in \Delta^{K-1}$ where $K$ denotes the number of regimes and $\Delta^{K-1}$ denotes the $(K{-}1)$-dimensional probability simplex, i.e., the set of $K$-dimensional non-negative vectors that sum to 1, which specify the prior probability of each regime; (3) an exit posterior $\gamma_{t-1,L} \in \Delta^{K-1}$, which is the posterior distribution over regimes at the final layer of the preceding step; and (4) a bridge matrix $J^{(c_{t-1},c)} \in \mathbb{R}_{\ge 0}^{K \times K}$, which propagates the exit posterior at layer $L$ of step $t{-}1$ to the regime distribution at layer $1$ of step $t$. The joint likelihood takes the form:
\begin{multline}
p(\hat{h}_{t,1:L} \mid c_t{=}c,\, \gamma_{t-1,L}) \\
  = \Biggl[\sum_{k=1}^{K}
      \biggl(\sum_{j=1}^{K} \gamma_{t-1,L}(j)\;J^{(c_{t-1},c)}_{j,k}\biggr)
      \;\cdot\; b^{(c)}_k(\hat{h}_{t,1})\Biggr] \\
  \cdot \prod_{\ell=2}^{L}
    \Biggl[\sum_{k=1}^{K}
      \pi^{(c)}_k \;\cdot\; b^{(c)}_k(\hat{h}_{t,\ell})\Biggr].
\label{eq:gmm_joint}
\end{multline}

Each regime density $b^{(c)}_k$ in Equation~\ref{eq:gmm_joint} is a multivariate Gaussian with a diagonal covariance matrix. The log-density decomposes into the squared Mahalanobis distance~\cite{bishop2006pattern,mahalanobis2018generalized}, a log-determinant term, and a normalization constant:
\begin{align}
\log b^{(c)}_k(x)
  &= -\frac{1}{2}\sum_{d=1}^{D}
      \frac{(x_d-\mu^{(c)}_{k,d})^2}{\sigma^{2(c)}_{k,d}} \nonumber\\
  &\quad -\frac{1}{2}\sum_{d=1}^{D}\log\sigma^{2(c)}_{k,d}
    - \frac{D}{2}\log(2\pi),
\label{eq:emission}
\end{align}
where $\mu^{(c)}_k \in \mathbb{R}^D$ is the regime mean vector and $\sigma^{2(c)}_k \in \mathbb{R}^D_{>0}$ the per-dimension variance vector.

\paragraph{Bridge matrix.}
\label{sec:bridge}
To model dependencies between consecutive steps, we introduce a \emph{bridge matrix} that links the exit regime of step $t-1$ to the entry regime of step $t$ and \emph{explicit stage} and the \emph{implicit stage}. We define the \emph{bridge matrix} $J^{(c,c')} \in \mathbb{R}_{\ge 0}^{K \times K}$ learned via EM, whose $(j,k)$-th entry is defined as $J^{(c,c')}_{j,k} = P(c_t{=}c',\, z_{t,1}{=}k \mid c_{t-1}{=}c,\, z_{t-1,L}{=}j)$, where $z_{t,\ell} \in \{1,\dots,K\}$ denotes the latent regime at layer $\ell$ of step $t$, representing the joint probability of transitioning to category $c'$ and entering latent regime $k$ at the first layer of step $t$, given that step $t-1$ ended in category $c$ with latent regime $j$ at its final layer.

We use $J$ replaces $\pi^{(c)}$ as the layer-$1$ prior (see Equation~\ref{eq:gmm_joint}), providing a context-dependent entry distribution that depends on the preceding step's exit regime $j$ and category transition $c\to c'$. We compute the full joint posterior as
$P(z_{t-1,L}{=}j,\, z_{t,1}{=}k \mid \hat{h}_{t-1,L}) \notag \quad\propto\; \gamma_{t-1,L}(j) \;\cdot\; J^{(c,c')}_{j,k} \;\cdot\; b^{(c')}_k(\hat{h}_{t,1})$,
where $\gamma_{t-1,L}(j)$ is the exit posterior of the preceding step, $J^{(c,c')}_{j,k}$ is the current bridge estimate, and $b^{(c')}_k(\hat{h}_{t,1})$ is the emission density at layer~1 of the current step; the result is normalized over all $(j,k)$ to yield a proper distribution. 

From $J$, we obtain the category transition probability conditioned on the exit regime by marginalizing over entry regimes:
$
R_{j,c,c'} = \sum_{k} J^{(c,c')}_{j,k}
= P(c_t = c' \mid c_{t-1} = c,\, z_{t-1,L} = j).
$
We refer to $R$ as the \emph{explicit bridge}, which captures how category transitions depend on the preceding step’s exit regime.

\paragraph{Training.}
Training proceeds in two phases. In Phase~1, the GMM parameters $\{\mu^{(c)}_k, \sigma^{(c)2}_k, \pi^{(c)}_k\}$ are trained via standard EM without bridge matrices for $n_{\mathrm{warmup}}$ iterations. This provides a warm start for the regime structure.
In Phase~2, we first initialize the bridge $J$ from the Phase~1 posteriors via outer products $\gamma_{t-1,L}(j) \cdot \gamma_{t,1}(k)$. We then run joint EM iterations where E-step uses $J$ for the first layer prior (Equation~\ref{eq:gmm_joint}), followed by M-steps that update parameters. Details on the EM procedure, decoding process, and the selection of the number of regimes $K$ are provided in Appendix~\ref{app:method}.
\section{Experimental Setup}
\label{sec:experiment}

\paragraph{Models.} 
We evaluate four open-source reasoning models spanning different parameter scales (1.7B--7B) and architectures: Bespoke-Stratos-7B \citep{bespoke_stratos}, OpenThinker-7B \citep{guha2025openthoughtsdatarecipesreasoning}, Qwen3-1.7B \citep{qwen3technicalreport}, and Llama-3.1-Nemotron-Nano-4B-v1.1 \citep{bercovich2025llama}. Bespoke-Stratos and OpenThinker are both distilled from DeepSeek-R1 on different datasets using Qwen2.5-7B-Instruct as the base.

\paragraph{Datasets.} 
Our experiments span four benchmarks from different domains: MATH-500 \citep{lightman2023lets} (competition mathematics), AIME-2024 \citep{huggingface2024aime} (competition mathematics), GPQA-Diamond \citep{rein2024gpqa} (graduate-level science), WebInstruct-Verified \citep{general-reasoner} (diverse real-world problems).
\section{PRISM Components and Analysis}
\label{sec:results}

Using PRISM, we characterize the structure of reasoning behavior by uncovering the semantic flows and latent computational patterns underlying each semantic category.

\subsection{Explicit Stage: Understanding Semantic Flow of Reasoning Trajectories}
\label{sec:results_explicit}

We first use PRISM to characterize the semantic flow of reasoning trajectories. Specifically, we analyze how reasoning steps move among the four core categories: SR (Setup \& Retrieval), AC (Analysis \& Computation), UV (Uncertainty \& Verification), and FA (Final Answer) while excluding \texttt{unknown}, which corresponds to ambiguous steps whose functional role is unclear. This view reveals how a reasoning trajectories flow.

\paragraph{Q1: How do reasoning trajectories flow?}

Reasoning does not proceed randomly. The first-order transition matrix (Table~\ref{tab:markov-summary}) reveals structured patterns along three dimensions:

\paragraph{Self-loops.} Reasoning is dominated by Analysis \& Computation self-transitions ($p_{\text{AC}\to\text{AC}} = 0.715$). Setup \& Retrieval steps most frequently transit into Analysis \& Computation ($p_{\text{SR}\to\text{AC}} = 0.445$), establishing a setup-to-computation pathway. The stationary distribution confirms that Analysis \& Computation dominates the long-run behavior ($\pi_{\text{AC}} = 0.501$), followed by Uncertainty \& Verification ($\pi_{\text{UV}} = 0.238$). We fit the reasoning traces into higher order and find the second order analysis has the lowest BIC score, see Appendix~\ref{app:method}. The most frequent two-step pathway is AC$\to$AC$\to$AC ($p = 0.762$), while the rarest is SR$\to$SR$\to$FA ($p = 0.015$) (full table see Appendix~\ref{app:results}), indicating that setup phases rarely transition directly to a final answer without an intermediate computation step.

\paragraph{Verification cycles.} Uncertainty \& Verification steps are most likely to loop back into Analysis \& Computation ($p_{\text{UV}\to\text{AC}} = 0.320$) or self-loop ($p_{\text{UV}\to\text{UV}} = 0.438$), and UV$\to$AC$\to$UV has a probability of 27.1\%, suggesting that verification often triggers further computation or continued checking.

\paragraph{Start and end.} Nearly all samples begin with Setup \& Retrieval ($p_{\text{start}} = 0.982$), indicating a near-universal setup phase before substantive reasoning. The expected hitting time from any non-Final Answer category to Final Answer is approximately 22 steps, suggesting that reaching a final answer requires sustained multi-step reasoning regardless of the current semantic role.

\begin{table}[t]
\centering
\resizebox{\columnwidth}{!}{%
\begin{tabular}{@{}l cccc@{}}
\toprule
\textbf{From $\downarrow$ / To $\rightarrow$}
& FA & SR & AC & UV \\
\midrule
FA & 0.466 & 0.076 & 0.131 & 0.326 \\
SR & 0.038 & 0.419 & 0.445 & 0.098 \\
AC & 0.060 & 0.079 & \textbf{0.715} & 0.147 \\
UV & 0.124 & 0.119 & \textbf{0.320} & \textbf{0.438} \\
\midrule
\textbf{Start Distribution}       & 0.000 & 0.982 & 0.007 & 0.011 \\
\textbf{Stationary Dist.}  & 0.128 & 0.133 & 0.501 & 0.238 \\
\textbf{Expected Steps to FA}     & - & 22.52 & 21.99 & 21.30 \\
\bottomrule
\end{tabular}%
}
\caption{First-order Markov chain summary (aggregated over all models and datasets). Top: transition matrix. Bottom: start distribution, stationary distribution, and expected steps to Final Answer. Categories: FA = Final Answer, SR = Setup \& Retrieval, AC = Analysis \& Computation, UV = Uncertainty \& Verification.}
\label{tab:markov-summary}
\end{table}

\subsection{Implicit Stage: Understanding Internal Computations}
\label{sec:results_implicit}

So far, we have examined \emph{what} the model says at each step. PRISM also lets us examine \emph{how} it says it: the internal computation across transformer layers. We use Bespoke-Stratos-7B on GPQA-Diamond benchmark as illustrative example to show how PRISM find its implicit structure.

\begin{figure}[h]
    \centering
    \includegraphics[width=\linewidth]{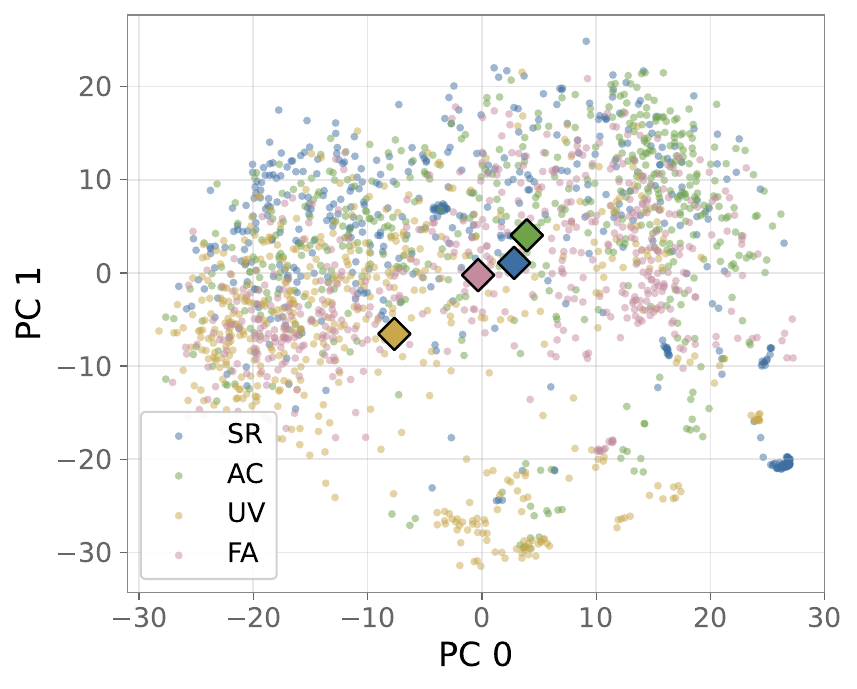}
    \caption{2D PCA projection of per-step activations colored by category. Diamond markers = category centroids. Separation between clusters indicates distinct hidden-state representations per category.}
    \label{fig:cas}
\end{figure}

\paragraph{Q2: What internal structure exists within semantic categories?}~\\
\emph{Category activations exhibit spatial separation in the activation space.} We project per-step activations onto the first two principal components (Figure~\ref{fig:cas}) and observe that, for this model and dataset, the categories occupy different regions of the PCA plane: UV appears separated (centroid at $(-7.66, -6.55)$), whereas SR, AC, and FA remain more tightly clustered.

\emph{The exit regime of a step can influence which category follows.} The explicit bridge
$R_{j,c,c'} = P(c_t{=}c' \mid c_{t-1}{=}c, z_{t-1,L}{=}j)$ models the dependency between the exit regime and the next category. Table~\ref{tab:bridge} shows that, in some cases, different exit regimes lead to different transition tendencies. For example, in this model and dataset, a Setup \& Retrieval step exiting through Regime 1 (R1 in Table~\ref{tab:bridge}) in the last layer is more likely to remain in Setup \& Retrieval ($p = 0.80$) than one exiting through R0 ($p = 0.35$). In contrast, exits through Regime 0 more often transition to Analysis \& Computation. This effect is category-specific. For instance, transitions following Analysis \& Computation remain relatively consistent across exit regimes, indicating that some categories exhibit stable transition dynamics, whereas others (e.g., Setup \& Retrieval in this case) show stronger regime-dependent behavior captured by the bridge.

\begin{table}[t]
\centering
\resizebox{\columnwidth}{!}{
\begin{tabular}{c cccc | c cccc}
\toprule
\cmidrule(lr){2-5} \cmidrule(lr){7-10}
From SR & FA & SR & AC & UV & From AC & FA & SR & AC & UV \\
\midrule
SR R0 & 5\% & \textbf{35\%} & \textbf{46\%} & 14\% & AC R0 & 0\% & 0\% & 87\% & 13\% \\
SR R1 & 0\% & \textbf{80\%} & 9\% & 11\% & AC R1 & 0\% & 0\% & 99\% & 1\% \\
SR R2 & 0\% & 0\% & 2\% & 98\% & AC R2 & 0\% & 0\% & 0\% & 100\% \\
SR R3 & 4\% & 69\% & 25\% & 3\% & AC R3 & 5\% & 12\% & 61\% & 22\% \\
SR R4 & 6\% & 22\% & 62\% & 9\% & AC R4 & 23\% & 13\% & 46\% & 18\% \\
SR R5 & 0\% & 99\% & 0\% & 0\% & AC R5 & 4\% & 8\% & 68\% & 19\% \\
\bottomrule
\end{tabular}
}
\caption{The Explicit bridge
$P(\text{target cat} \mid \text{source cat},\, \text{exit regime})$ 
(Stratos, GPQA-Diamond, $K{=}6$).}
\label{tab:bridge}
\end{table}

\section{Application with PRISM: Failure Patterns \& Overthinking}
Many reasoning phenomena have been observed in LLM reasoning trajectories, including overthinking~\cite{sui2025stop}, where models generate long reasoning chains, and premature commitment, where models prematurely terminate reasoning. PRISM provides an informative view of such failures than aggregate accuracy alone.

\begin{table*}[h]
\centering
\resizebox{\textwidth}{!}{%
\begin{tabular}{@{}l rrrr rrrr rrrr@{}}
\toprule
& \multicolumn{4}{c}{\textbf{Correct $-$ Incorrect}}
& \multicolumn{4}{c}{\textbf{Correct $-$ Long Fail}}
& \multicolumn{4}{c}{\textbf{Correct $-$ Short Fail}} \\
\cmidrule(lr){2-5} \cmidrule(lr){6-9} \cmidrule(l){10-13}
\textbf{From / To}
& FA & SR & AC & UV
& FA & SR & AC & UV
& FA & SR & AC & UV \\
\midrule
FA
  & $-$3.4\% & +2.2\% & 0.0\% & +1.2\%
  & $-$1.2\% & +2.3\% & $-$2.2\% & +1.1\%
  & $-$0.7\% & $-$2.6\% & +3.0\% & +0.2\% \\
SR
  & $-$0.9\% & $-$0.9\% & +5.1\% & $-$3.3\%
  & $-$0.6\% & +0.6\% & +3.4\% & $-$3.5\%
  & $-$1.7\% & $-$5.3\% & +11.4\% & $-$4.3\% \\
AC
  & +0.4\% & $-$1.5\% & +5.1\% & \textbf{$-$4.1\%}
  & \textbf{+1.4\%} & $-$1.6\% & +4.3\% & $-$4.0\%
  & \textbf{$-$3.7\%} & $-$2.4\% & +14.0\% & $-$8.0\% \\
UV
  & $-$0.9\% & $-$0.5\% & \textbf{+5.9\%} & $-$4.5\%
  & $-$0.6\% & $-$0.6\% & +5.6\% & $-$4.4\%
  & +1.7\% & $-$0.7\% & +5.3\% & $-$6.3\% \\
\bottomrule
\end{tabular}%
}
\caption{Semantic transition matrix differences ($\Delta P = P_{\text{correct}} - P_{\text{fail}}$). Positive values indicate higher transition probability in correct reasoning. Computed by pooling all model-dataset configurations, standard deviations are in Appendix~\ref{app:results}.}
\label{tab:trans-diff}
\end{table*}

\paragraph{Correct and incorrect reasoning follow different semantic transition paths.} 
As shown in Table~\ref{tab:trans-diff}, incorrect samples transition more readily from Analysis \& Computation to Uncertainty \& Verification ($-0.041$) yet less frequently return from Uncertainty \& Verification to Analysis \& Computation ($+0.059$). This indicates that failed reasoning tends to enter verification but struggle to exit, forming unproductive checking loops, whereas correct reasoning proceeds with more efficient computation. 

To analyze differences among failed trajectories, we partition them into two groups using a path-length threshold of 100 steps: \emph{long-fail} ($\geq 100$ steps; $50.7\%$) and \emph{short-fail} ($< 100$ steps; $49.3\%$). The two groups exhibit different behaviors around the Final Answer stage. In the transition statistics, long failures show slightly reduced AC$\to$FA flow ($\Delta p = +0.014$), whereas short failures show elevated AC$\to$FA flow ($\Delta p = -0.037$), suggesting that short failures exhibit earlier commitment from analysis to final answer. Table~\ref{tab:overthinking} further examines how trajectories interact with the Final Answer state. Long failures reach Final Answer earlier (median relative position $0.28$) and repeatedly re-enter it ($8$). After the first FA visit, $90\%$ of subsequent steps are non-FA, indicating continued deliberation after proposing a candidate answer. Short failures instead reach Final Answer later ($0.58$) and re-enter it less frequently ($3$), suggesting that the model tends to stop reasoning shortly after an answer appears. These Final Answer dynamics reveal two different behaviors: repeated revisiting of candidate answers, consistent with \emph{overthinking}, and early commitment to an answer, consistent with \emph{premature commitment}.

\begin{figure}[t]
    \centering
    \includegraphics[width=\linewidth]{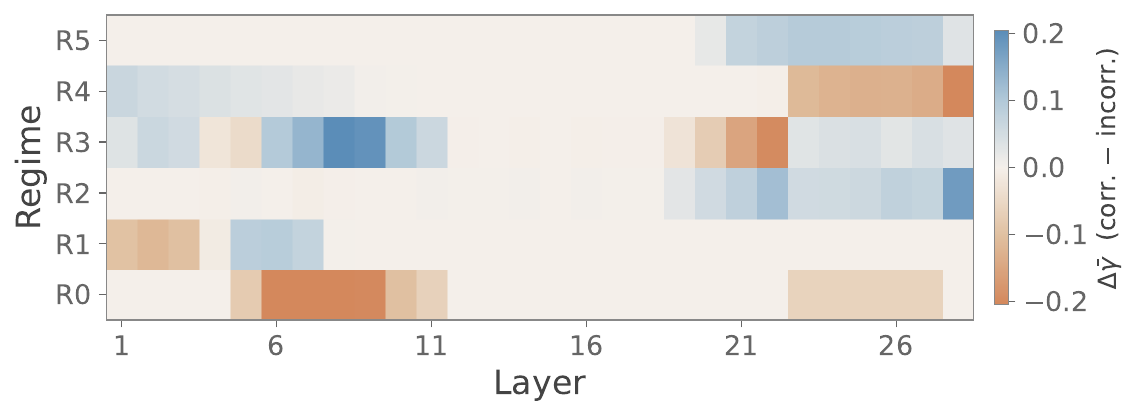}
    \caption{Mean posterior probability $\gamma_{\ell,k}$ difference averaged across all steps (blue = more in correct, red = more in incorrect). Columns = transformer layers (left=early, right=late), rows = regimes. Strong colored cells in the difference plot indicate (layer, regime) pairs that discriminate correct from incorrect reasoning.}
    \label{fig:scp}
\end{figure}

\begin{figure}[h]
    \centering
    \includegraphics[width=\linewidth]{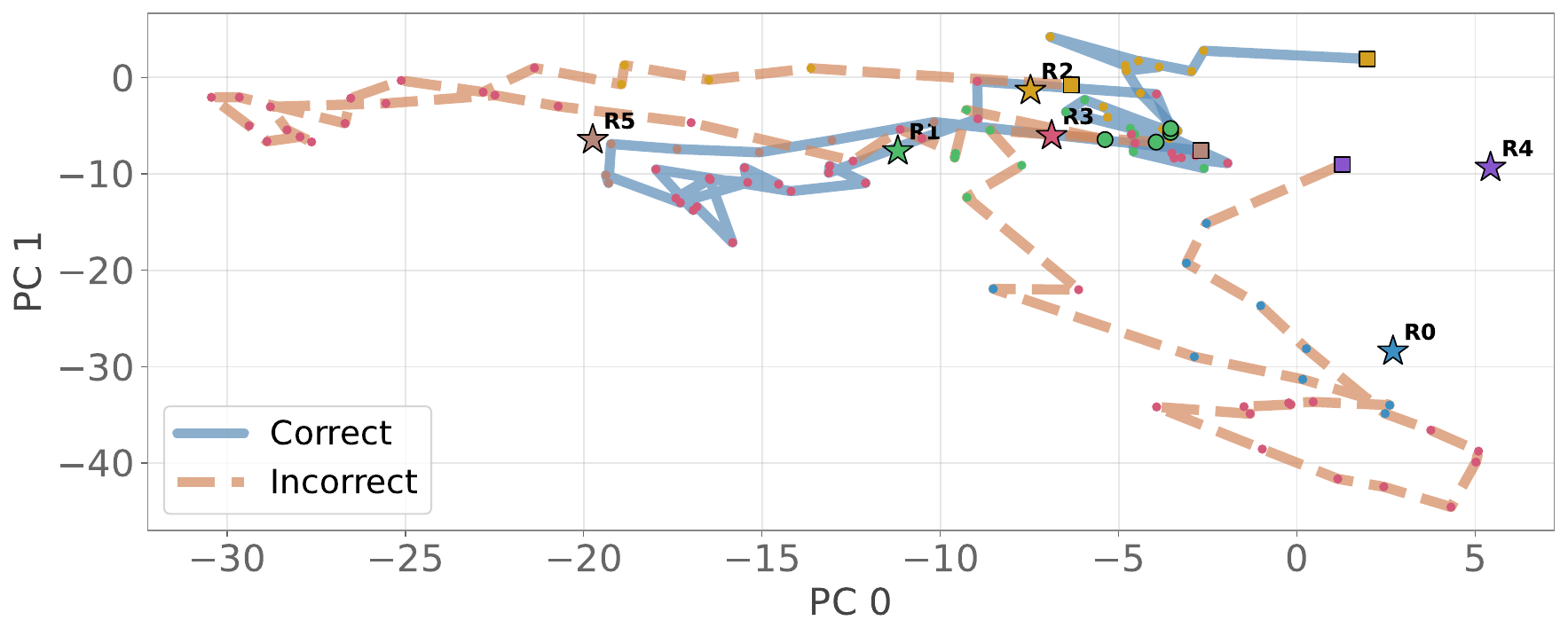}
    \caption{Step-level layer trajectories in PCA space (Stratos, GPQA-Diamond, UV). Each path traces the hidden-state path of a single reasoning step across transformer layers ($\ell = 1 \to L$) projected onto the first two PCA dimensions. Points are colored by their decoded regime ($R_0$--$R_{K-1}$). Stars mark per-regime centroids. Solid blue paths correspond to correct sequences; dashed orange paths to incorrect sequences.}
    \label{fig:st}
\end{figure}

\paragraph{Correct and incorrect reasoning engage different internal structures.} 
We compute soft posteriors $\gamma_{\ell,k} = P(z_\ell = k \mid \hat{h}_{t,\ell})$. The mean posterior $\bar{\gamma}_{\ell,k}$, averaged across all steps of a given category, highlights regimes that discriminate between correct and incorrect reasoning. Taking Uncertainty \& Verification of Bespoke-Stratos-7B on GPQA-Diamond as an example (Figure~\ref{fig:scp}), Regime~4 in the final layer appears more frequently in incorrect samples ($\Delta\bar{\gamma}_{28,4} \approx -0.24$), Regime~0 shows more in earlier layers (layers~6--9). This suggests that incorrect UV steps engage different computational modes in the early and late transformer layers, with little difference observed in the middle layers. For clearer visualization, we plot the step trajectory for some random samples. From Figure~\ref{fig:st}, we can see it differ in which regimes dominate at early layers and only incorrect path shows regime 0 in early layers.

\begin{table*}[t]
\centering
\resizebox{\textwidth}{!}{
\begin{tabular}{ll*{4}{c}*{4}{c}*{4}{c}*{3}{c}|c}
\toprule
& & \multicolumn{4}{c}{Stratos} & \multicolumn{4}{c}{OpenThinker} & \multicolumn{4}{c}{Qwen} & \multicolumn{3}{c|}{Nemotron} & \\
\cmidrule(lr){3-6} \cmidrule(lr){7-10} \cmidrule(lr){11-14} \cmidrule(lr){15-17}
& & MATH500 & GPQA-D & AIME24 & WebInstruct
& MATH500 & GPQA-D & AIME24 & WebInstruct
& MATH500 & GPQA-D & AIME24 & WebInstruct
& MATH500 & GPQA-D & WebInstruct & All \\
\midrule
\multicolumn{18}{l}{\textit{Implicit Stage $L_2$ divergence (long fail vs.\ short fail)}} \\
& FA & \textbf{2.36} & 0.91 & 1.31 & \textbf{2.05} & \textbf{4.63} & \textbf{2.53} & \textbf{3.30} & \textbf{5.25} & \textbf{1.52} & 0.65 & \textbf{2.40} & 1.04 & \textbf{3.15} & 1.47 & 1.46 & \textbf{2.27} \\
& SR & 1.19 & \textbf{2.06} & \textbf{1.56} & 1.30 & 2.31 & 0.74 & 1.14 & 1.74 & 1.37 & \textbf{0.93} & 2.21 & \textbf{1.14} & 1.99 & 1.38 & \textbf{1.75} & 1.52 \\
& AC & 0.44 & 1.34 & 1.10 & 0.65 & 0.61 & 1.45 & 0.89 & 0.75 & 0.92 & 0.73 & 2.16 & 0.49 & 1.09 & \textbf{1.71} & 0.73 & 1.00 \\
& UV & 1.41 & 1.10 & 1.51 & 1.55 & 1.35 & 2.34 & 1.63 & 1.14 & 1.13 & 0.53 & 1.03 & 1.01 & 1.92 & 0.55 & 0.90 & 1.27 \\
\midrule
\multicolumn{18}{l}{\textit{FA visit metrics (median)}} \\
\multirow{2}{*}{\shortstack[l]{FA\\pos}}
& long & 0.29 & 0.25 & 0.36 & 0.21 & 0.28 & 0.30 & 0.33 & 0.24 & 0.37 & 0.33 & 0.61 & 0.23 & 0.29 & 0.46 & 0.27 & 0.28 \\
& short & 0.72 & 0.45 & 0.44 & 0.59 & 0.59 & 0.59 & 0.74 & 0.53 & 0.37 & 0.55 & 0.50 & 0.57 & 0.73 & 0.69 & 0.73 & 0.58 \\
\midrule
\multirow{2}{*}{\shortstack[l]{FA\\re-enter}}
& long & 7 & 7 & 5 & 10 & 8 & 7 & 7 & 10 & 8 & 7 & 8 & 13 & 10 & 5 & 13 & 8 \\
& short & 2 & 3 & 7 & 3 & 3 & 3 & 4 & 3 & 4 & 4 & 4 & 3 & 2 & 2 & 2 & 3 \\
\midrule
\multirow{2}{*}{\shortstack[l]{Post\\FA}}
& long & 0.95 & 0.89 & 0.97 & 0.90 & 0.93 & 0.85 & 0.96 & 0.88 & 0.91 & 0.86 & 0.89 & 0.87 & 0.92 & 0.90 & 0.88 & 0.90 \\
& short & 0.67 & 0.73 & 0.80 & 0.67 & 0.70 & 0.71 & 0.62 & 0.71 & 0.81 & 0.77 & 0.81 & 0.62 & 0.60 & 0.75 & 0.62 & 0.71 \\
\bottomrule
\end{tabular}
}
\caption{Overthinking analysis per model--dataset configuration. Top: implicit stage $L_2$ divergence between long and short failures (bold = highest category). Bottom: FA visit metrics (median values). ``FA pos'' = relative position of first FA visit; ``FA re-enter'' = number of contiguous FA blocks; ``Post-FA'' = fraction of steps after the first FA that are non-FA.}
\label{tab:overthinking}
\end{table*}

Table~\ref{tab:overthinking} compares the implicit stage mean posterior probability profiles of long and short fail.  Final Answer and Setup \& Retrieval show the larger divergence across model–dataset combinations, whereas Analysis \& Computation and Uncertainty \& Verification exhibit smaller differences. This suggests that the two failure modes engage qualitatively different computational regimes during problem setup and answer formulation.
\section{Application with PRISM: Prompting Strategy Reshape Reasoning Flow}
\label{sec:results_prompt}

\begin{table*}[t]
\centering
\resizebox{\textwidth}{!}{%
\begin{tabular}{@{}l cccc cccc cccc | ccc@{}}
\toprule
& \multicolumn{12}{c}{\textbf{1st-order TransMat} $P(j \mid i)$}
& \multicolumn{3}{c}{\textbf{Implicit $L_2$ vs.\ P1}} \\
\cmidrule(lr){2-13} \cmidrule(l){14-16}
\textbf{Category}
& \multicolumn{4}{c}{\textbf{P1 (Baseline)}}
& \multicolumn{4}{c}{\textbf{P2 (Concise)}}
& \multicolumn{4}{c}{\textbf{P3 (Explore)}}
& \textbf{Category} & P2 & P3 \\
\cmidrule(lr){2-5} \cmidrule(lr){6-9} \cmidrule(l){10-13}
& FA & SR & AC & UV
& FA & SR & AC & UV
& FA & SR & AC & UV
& & & \\
\midrule
FA
  & 0.444 & 0.068 & 0.142 & 0.345
  & 0.464 & 0.069 & 0.169 & 0.299
  & 0.411 & 0.085 & 0.184 & 0.320
  & FA & 0.265 & 0.272 \\
SR
  & 0.036 & 0.383 & 0.434 & 0.147
  & 0.043 & 0.372 & 0.465 & 0.120
  & 0.038 & 0.400 & 0.424 & 0.137
  & SR & 0.173 & 0.218 \\
AC
  & 0.055 & 0.104 & 0.636 & 0.205
  & 0.052 & 0.115 & 0.549 & 0.284
  & 0.042 & 0.091 & 0.536 & 0.331
  & AC & 0.193 & 0.483 \\
UV
  & 0.050 & 0.141 & 0.270 & 0.539
  & 0.051 & 0.150 & 0.284 & 0.515
  & 0.045 & 0.109 & 0.373 & 0.473
  & UV & 0.292 & 0.339 \\
\midrule
\textbf{Stationary $\pi$}
  & 0.083 & 0.156 & 0.450 & 0.312
  & 0.085 & 0.165 & 0.414 & 0.336
  & 0.068 & 0.140 & 0.439 & 0.353
  & -- & -- & -- \\
$\mathbb{E}[\text{steps to FA}]$
  & 0.0 & 20.2 & 19.7 & 19.9
  & 0.0 & 20.3 & 20.1 & 20.1
  & 0.0 & 23.6 & 23.4 & 23.4
  & -- & -- & -- \\
\midrule
\textbf{Accuracy}
  & \multicolumn{4}{c}{35.9\%}
  & \multicolumn{4}{c}{38.9\%}
  & \multicolumn{4}{c}{37.9\%}
  & \multicolumn{3}{c}{--} \\
\textbf{Avg Tokens / Steps}
  & \multicolumn{4}{c}{5129.7 / 149.6}
  & \multicolumn{4}{c}{4286.0 / 120.5}
  & \multicolumn{4}{c}{6101.1 / 160.2}
  & \multicolumn{3}{c}{--} \\
\bottomrule
\end{tabular}%
}
\caption{Comparison of prompting strategies (Stratos, GPQA-Diamond, seed 42). For each prompt, the top rows report first-order transition matrices $P(j \mid i)$. The rightmost block reports the implicit stage regime activation profile $L2$ distance relative to P1 (Baseline). Bottom rows summarize the stationary distribution, expected hitting time to FA, and overall performance statistics.}
\label{tab:prompt_results}
\end{table*}

Prompting and intervention strategies are often evaluated only through aggregate outcomes, which gives limited insight into how they actually change reasoning behavior. Here PRISM provides a more informative view: rather than solely evaluate on only whether a prompt improves accuracy, PRISM can also evaluate which parts of the reasoning process it reshapes.

We select three prompting strategies that target different reasoning behavior. \textit{P1} is our baseline prompt (full text in Appendix~\ref{app:method}). \textit{P2} augments P1 with \textit{``Be concise.''}, testing whether brevity constraints compress or change the reasoning structure. \textit{P3} augments P1 with \textit{``First explore several independent reasoning paths to solve the problem. Write each reasoning path separately (e.g., Path 1, Path 2, Path 3), then decide which option is most supported.''}, encouraging explicit exploration of multiple solution paths before committing to an answer. Results see Table~\ref{tab:prompt_results}.

\emph{P2 (be concise) fragments reasoning.} The conciseness instruction reduces the Analysis \& Computation self-loop ($0.549$ vs.\ $0.636$) while increasing Analysis \& Computation $\to$ Uncertainty \& Verification transitions ($0.284$ vs.\ $0.205$). The stationary probability of Analysis \& Computation remains similar to P1. Taking together, this means computation is broken into shorter segments that alternate more often with verification, rather than forming long continuous computation blocks.

\emph{P3 (multi-path exploration) promotes verification cycling.} The exploration instruction yields the highest Uncertainty \& Verification $\to$ Analysis \& Computation transition ($0.331$ vs.\ $0.205$) and the highest Analysis \& Computation $\to$ Uncertainty \& Verification transition ($0.374$ vs.\ $0.270$), while producing the lowest Uncertainty \& Verification self-loop ($0.473$). At the implicit stage, P3 Analysis \& Computation ($0.483$) and Uncertainty \& Verification ($0.339$) shows largest divergence from P1. These patterns indicate stronger bidirectional cycling between computation and verification. This behavior is consistent with the intended multi-path exploration: the model alternates between developing a reasoning path and evaluating it, rather than remaining trapped in unproductive self-loops.
\section{Related Work}

Recent studies highlight that reasoning models can exhibit unintended or deceptive behaviors, underscoring the need for a deeper mechanistic understanding \citep{baker2025monitoring}. Several works examine which aspects of CoT steps matter. \citet{madaan2022text} disentangle the roles of textual content and structural patterns, \citet{wang2022towards} show that performance gains often persist even with flawed step content, and \citet{bogdan2025thought} find that a small set of “anchor” steps disproportionately shape final outcomes.

Mechanistic analyses probe the internal processes behind reasoning. \citet{cabannes2024iteration} and \citet{dutta2024think} show how architectural components enable stepwise reasoning. Latent multi-hop phenomena are revealed by \citet{yang2024large} and \citet{shalev2024distributional}, while \citet{venhoff2025understanding} demonstrate that steering vectors can capture functional directions in hidden space. Information-theoretic perspectives provide a complementary lens: \citet{Ton2024UnderstandingCI} analyze CoT dynamics through entropy and mutual information, while \citet{Punjwani2025WeightofThoughtRE} explore how neural network weights encode and constrain reasoning capacity.
\citet{hu2026towards} provides a taxonomy of reasoning failures. Reasoning efficiency survey \citep{sui2025stop} catalog methods to mitigate “overthinking,” and token-level analyses \citep{wang2025beyond} identify sparse high-entropy tokens as critical intervention points. \citet{yang2025understanding} links observable “aha” reasoning behaviors to underlying internal mechanisms. 
\citet{jiang2025makes} analyzes how structural patterns in reasoning trajectories relate to performance. \citet{wu2025ctrls} frame CoT as a latent-state MDP, training a transition policy with RL to improve reasoning exploration. \citet{yu2025explainable} cluster final-layer embeddings of steps and construct a Markov chain to visualize reasoning motifs. \citet{li2026interpreting} introduces a method to localize and steer reasoning behavior in LLMs by linking final outcomes to intermediate computations via policy-gradient-style attribution.
Complementary to these lines of work, PRISM introduces a unified probabilistic framework that connects semantic transitions with latent computational patterns, enabling structured and interpretable analysis of reasoning trajectories.
\section{Conclusion}

We introduced \textit{PRISM} (\textbf{P}robabilistic \textbf{R}easoning \textbf{I}nspection through \textbf{S}emantic and \textbf{I}mplicit \textbf{M}odeling), a framework for analyzing reasoning processes in LLMs by jointly modeling their semantic reasoning trajectories and internal computational structures. By combining an explicit stage that captures semantic category transitions with an implicit stage that identifies latent computational regimes across transformer layers, PRISM provides a unified view that links what a reasoning step does with how it is realized inside the network. 

PRISM enables practical diagnostic applications. It allows us to analyze the dynamics of correct and incorrect reasoning trajectories and uncover systematic patterns, including distinct behaviors across different failure modes. In addition, PRISM supports the analysis of prompting beyond final accuracy, revealing how prompt variations reshape both semantic transitions and internal computational patterns, which in turn influence end-task performance. Overall, PRISM provides a probabilistic framework for studying reasoning dynamics in language models, connecting internal computation with external reasoning structure. This perspective enables more interpretable reasoning systems and offers tools for analyzing and diagnosing the reasoning processes of language models.

\section*{Limitations}
PRISM relies on external semantic labeling and unsupervised latent clustering, which may introduce noise or ambiguity in category assignments and regime interpretation. In addition, our evaluation is conducted on a limited set of models and benchmarks, and further validation on broader architectures and tasks would strengthen the generality of our findings. Finally, PRISM is primarily designed as an analysis and diagnostic framework for analyzing reasoning dynamics and does not directly optimize or improve reasoning performance, although the insights it provides may inform future optimization.

\bibliography{custom}

\clearpage
\appendix
\section{Experimental Setup}
\label{app:method}

\subsection{Models}

We evaluate four open-source reasoning models. Licenses and key parameters are listed below.
 
\emph{Qwen3-1.7B} \citep{qwen3technicalreport} is used with ``thinking mode'' enabled. The parameters are: temperature = 0.6, top\_p = 1.0, top\_k = 20, and min\_p = 0. License: Apache 2.0. 

\emph{Bespoke-Stratos-7B} \citep{bespoke_stratos} is used with ``You are a reasoning assistant." system prompt. The configuration used temperature = 0.6, top\_p = 0.95. License: Apache 2.0. 

\emph{OpenThinker-7B} \citep{guha2025openthoughtsdatarecipesreasoning}  is used with ``You are a reasoning assistant." system prompt. The configuration used temperature = 0.6, top\_p = 0.95. License: Apache 2.0.  

\emph{Llama-3.1-Nemotron-Nano-4B-v1.1} \citep{bercovich2025llama} is used with ``Thinking mode'' enabled. The parameters are temperature = 0.6, top\_p = 0.95. License: NVIDIA Open Model License. 

\subsection{Datasets}

Our experiments span 4 benchmarks. Splits, sizes, maximum token and licenses are shown in Table~\ref{tab:datasets}.

\begin{table}[h]
\centering
\resizebox{\linewidth}{!}{
\begin{tabular}{l c c c c c}
\toprule
\textbf{Dataset} & \textbf{Split} & \textbf{Size} & \textbf{Max Tokens} & \textbf{License} \\
\midrule
MATH-500 \citep{lightman2023lets} & test & 500 & 32768 & MIT \\
GPQA-Diamond \citep{rein2024gpqa} & test & 198 & 32768 & Apache 2.0 \\
WebInstruct-Verified \citep{general-reasoner} & test & 288 & 32768 & Apache 2.0 \\
AIME-2024 \citep{huggingface2024aime} & train & 30 & 32768 & Apache 2.0 \\
\bottomrule
\end{tabular}}
\caption{Datasets used in experiments.}
\label{tab:datasets}
\end{table}

For WebInstruct-Verified, we filter by: answer\_types as Float, Multiple Choice, Integer, Percentage and difficulties as Primary, Junior High, Senior High.

\subsection{Prompts}

GPQA use:  
``
You are answering a multiple-choice question.
Options are labeled A, B, C, and D.
Think step-by-step and show your reasoning.
At the very end, output ONE line exactly in this format:
Final Answer: \boxed{A}
''

MATH-500, AIME-2024 and WebInstruct use:  
``
Answer the following question step-by-step. At the very end, output exactly one line formatted as:
Final Answer: \boxed{...}
''

\

\subsection{Semantic Category Classification}

Each step is assigned a semantic category using 
Qwen2.5-32B-Instruct~\cite{qwen2,qwen2.5} with the following prompt:

\begin{quote}
\small
Classify the sentence into ONE of these 4 tags:

- setup\_and\_retrieval: restates the problem or recalls known facts from the question \\
Example: ``The question is to find the value of $x$ such that $2x + 3 = 7$.'' \\
Example: ``Recall that the sum of angles in a triangle is 180 degrees.''

- analysis\_and\_computation: performs math, logic, or derivation \\
Example: ``Subtracting 3 from both sides gives $2x = 4$.'' \\
Example: ``Since $a + b = 10$ and $a = 3$, we have $b = 7$.''

- uncertainty\_and\_verification: expresses doubt or checks results \\
Example: ``Let me verify this by substituting back into the original equation.'' \\
Example: ``Wait, I think I made an error in the previous step.''

- final\_answer: states the final conclusion \\
Example: ``Therefore, the answer is 42.'' \\
Example: ``The final answer is $x = 2$.''

Output: \textbackslash\textbackslash boxed\{tag\_name\}
\end{quote}

\begin{table*}[t]
\centering
\resizebox{\textwidth}{!}{
\begin{tabular}{llcccccc|ccccc}
\toprule
Model & Dataset & $N$ & Acc & AvgTok & AvgStep & LongFail & ShortFail 
& FA$_n$ & SR$_n$ & AC$_n$ & UV$_n$ & UNK$_n$ \\
\midrule

Stratos & AIME24 & 30 & $14.4\%\pm1.9\%$ & $15122\pm1682$ & $369.3\pm50.3$ & $24.3\pm0.6$ & $1.3\pm0.6$ 
& $242\pm57$ & $1519\pm259$ & $6938\pm1502$ & $2375\pm310$ & $4\pm2$ \\

Stratos & GPQA Diamond & 198 & $39.9\%\pm4.0\%$ & $5091\pm155$ & $148.3\pm11.8$ & $58.3\pm3.8$ & $60.7\pm5.5$ 
& $2236\pm119$ & $6289\pm2457$ & $12301\pm950$ & $8532\pm594$ & $11\pm2$ \\

Stratos & MATH500 & 500 & $76.9\%\pm1.4\%$ & $4214\pm38$ & $105.4\pm2.6$ & $67.0\pm7.2$ & $48.3\pm5.0$ 
& $2840\pm156$ & $6688\pm421$ & $32676\pm726$ & $10489\pm457$ & $14\pm8$ \\

Stratos & WebInstruct & 288 & $43.5\%\pm2.3\%$ & $3516\pm148$ & $98.9\pm7.2$ & $53.0\pm5.6$ & $109.7\pm6.7$ 
& $3221\pm402$ & $3945\pm364$ & $15010\pm803$ & $6309\pm1117$ & $8\pm6$ \\

\midrule

OpenThinker & AIME24 & 30 & $34.4\%\pm5.1\%$ & $14855\pm1253$ & $410.0\pm65.2$ & $19.0\pm2.0$ & $0.7\pm0.6$ 
& $1043\pm1010$ & $1474\pm83$ & $7952\pm1181$ & $1827\pm37$ & $4\pm2$ \\

OpenThinker & GPQA Diamond & 198 & $42.8\%\pm2.1\%$ & $9328\pm340$ & $248.7\pm24.6$ & $62.3\pm2.1$ & $51.0\pm2.6$ 
& $19220\pm4806$ & $4669\pm63$ & $13172\pm632$ & $12176\pm813$ & $16\pm4$ \\

OpenThinker & MATH500 & 500 & $80.1\%\pm0.7\%$ & $4621\pm55$ & $122.8\pm4.0$ & $57.7\pm3.2$ & $41.7\pm2.5$ 
& $12257\pm739$ & $7501\pm243$ & $31612\pm1577$ & $10010\pm879$ & $9\pm4$ \\

OpenThinker & WebInstruct & 288 & $43.4\%\pm1.0\%$ & $5837\pm279$ & $148.9\pm23.7$ & $58.0\pm3.6$ & $105.0\pm4.0$ 
& $15612\pm6662$ & $4336\pm297$ & $16155\pm591$ & $6785\pm583$ & $6\pm1$ \\

\midrule

Qwen & AIME24 & 30 & $45.6\%\pm3.8\%$ & $14999\pm700$ & $338.3\pm25.0$ & $15.7\pm1.5$ & $0.7\pm0.6$ 
& $358\pm22$ & $1771\pm106$ & $5623\pm586$ & $2393\pm89$ & $6\pm1$ \\

Qwen & GPQA Diamond & 198 & $37.7\%\pm2.0\%$ & $7556\pm92$ & $135.6\pm6.3$ & $74.7\pm3.5$ & $48.7\pm2.3$ 
& $2308\pm65$ & $6303\pm288$ & $9675\pm556$ & $8544\pm382$ & $13\pm3$ \\

Qwen & MATH500 & 500 & $83.5\%\pm0.5\%$ & $4327\pm49$ & $107.7\pm3.2$ & $45.7\pm3.8$ & $36.7\pm2.1$ 
& $4298\pm151$ & $11288\pm388$ & $27907\pm846$ & $10353\pm387$ & $9\pm5$ \\

Qwen & WebInstruct & 288 & $46.3\%\pm2.3\%$ & $5218\pm56$ & $124.2\pm3.2$ & $70.0\pm7.0$ & $84.7\pm3.1$ 
& $3928\pm78$ & $5289\pm68$ & $17928\pm615$ & $8612\pm221$ & $2\pm1$ \\

\midrule

Nemotron & AIME24 & 30 & $62.2\%\pm1.9\%$ & $16539\pm820$ & $399.9\pm19.9$ & $11.3\pm0.6$ & $0.0\pm0.0$ 
& $398\pm64$ & $1340\pm169$ & $7838\pm612$ & $2417\pm56$ & $3\pm3$ \\

Nemotron & GPQA Diamond & 198 & $53.0\%\pm2.2\%$ & $12835\pm156$ & $189.6\pm4.5$ & $65.3\pm6.4$ & $27.7\pm2.1$ 
& $2207\pm457$ & $4318\pm325$ & $17049\pm266$ & $13954\pm686$ & $12\pm2$ \\

Nemotron & MATH500 & 500 & $87.2\%\pm0.2\%$ & $4326\pm127$ & $97.3\pm4.1$ & $29.0\pm1.0$ & $35.0\pm0.0$ 
& $3441\pm554$ & $5856\pm167$ & $29376\pm331$ & $9988\pm987$ & $11\pm2$ \\

Nemotron & WebInstruct & 288 & $42.2\%\pm1.3\%$ & $7440\pm165$ & $159.1\pm8.8$ & $76.3\pm3.1$ & $90.0\pm6.2$ 
& $5390\pm1332$ & $4041\pm138$ & $23561\pm1008$ & $12831\pm270$ & $9\pm8$ \\

\bottomrule
\end{tabular}
}
\caption{Dataset statistics and reasoning category counts across models. Values are mean $\pm$ std over three seeds.}
\label{tab:data-summary}
\end{table*}

\subsection{Preprocessing}

The raw hidden-state matrix $H \in \mathbb{R}^{L \times T \times d}$ is normalized and projected using PCA before being passed to the model.

\textit{Normalization.}
Each layer $\ell$ is mean-centered and scalar-RMS-normalized across all training steps:
\begin{equation}
\tilde{h}_{t,\ell} = \frac{h_{t,\ell} - \mu_\ell}{\rho_\ell},
\label{eq:layer_norm}
\end{equation}
where $\mu_\ell = \mathbb{E}_t[h_{t,\ell}]$ and $\rho_\ell = \sqrt{\mathbb{E}_{t}[\|h_{t,\ell}-\mu_\ell\|^2]/d}$ are estimated from training steps. This removes layer-specific biases and scale differences, ensuring that observations from all layers are comparable under a shared GMM.

All $L$ layer vectors of step $t$ are then divided by a single scalar to equalize activation magnitudes across steps while preserving every inter-layer relationship within a step:
\begin{equation}
\check{h}_{t,\ell} = \frac{\tilde{h}_{t,\ell}}{\rho_t^{\mathrm{step}}}, \qquad
\rho_t^{\mathrm{step}} = \sqrt{\frac{1}{Ld}\sum_{\ell=1}^{L}\|\tilde{h}_{t,\ell}\|^2}.
\label{eq:step_rms}
\end{equation}

\begin{figure}[h]
    \centering
    \includegraphics[width=\linewidth]{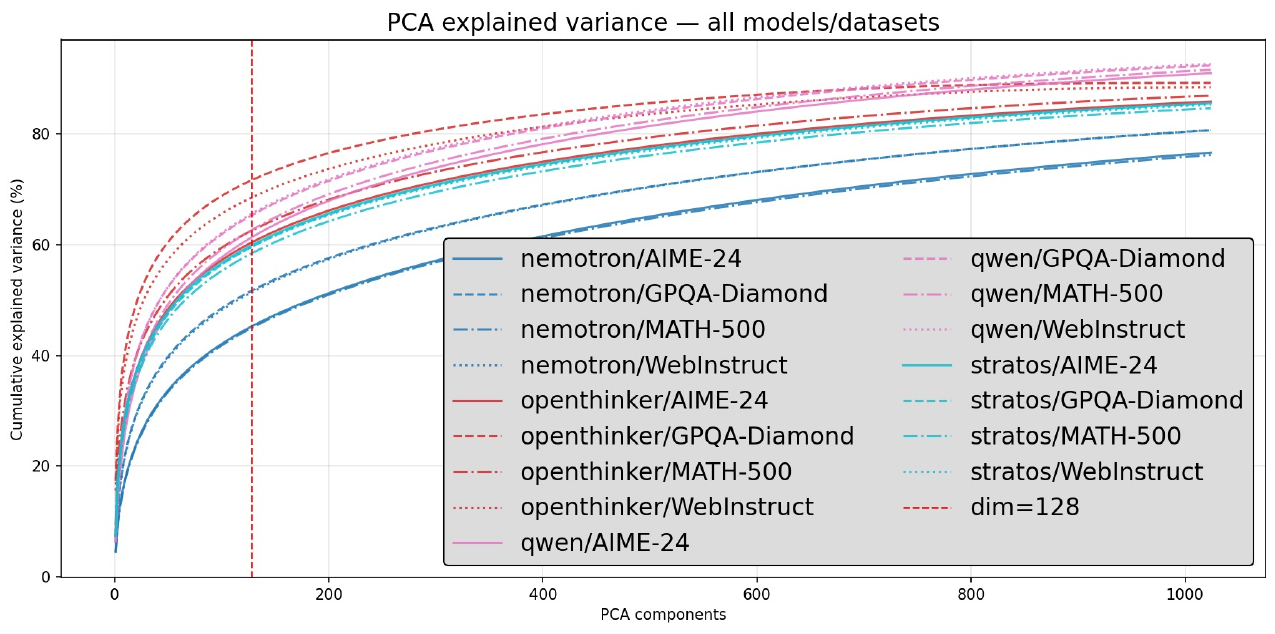}
    \caption{The cumulative explained variance curve tends to flatten after 128 dimensions, with smaller gains from additional components.}
    \label{fig:pca_curve}
\end{figure}

\textit{PCA.}
We fit a PCA on all training step-layer vectors and project to $d_{\mathrm{pca}}=128$ dimensions:
\begin{equation}
\hat{h}_{t,\ell}
  = W_{\mathrm{pca}}^\top\!\left(\check{h}_{t,\ell} - \bar{h}\right)
  \;\in\; \mathbb{R}^{d_{\mathrm{pca}}},
\label{eq:pca}
\end{equation}
where $W_{\mathrm{pca}}\in\mathbb{R}^{d\times d_{\mathrm{pca}}}$ are the top-$d_{\mathrm{pca}}$ principal components and $\bar{h}$ the global mean of $\check{h}_{t,\ell}$. We shows the explained variance curves in Figure~\ref{fig:pca_curve}.

\begin{figure}
    \centering
    \includegraphics[width=\linewidth]{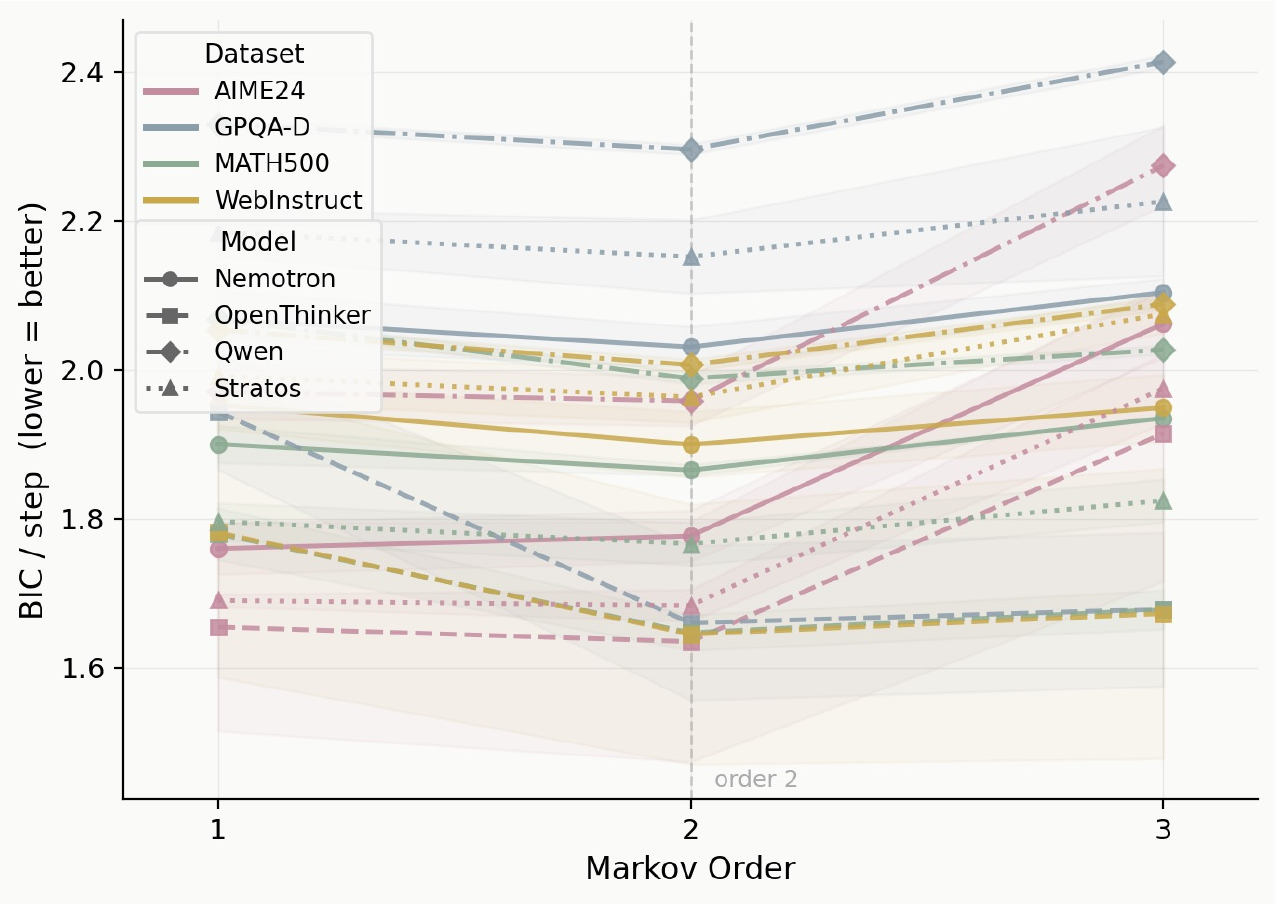}
    \caption{For the explicit stage, the second-order model achieves the lowest BIC score. Nemotron on AIME24 has relatively smaller BIC score on the 1st order.}
    \label{fig:placeholder}
\end{figure}

\subsection{EM Step}
\paragraph{E-step.}
Let $z_{t,\ell} \in \{1,\dots,K\}$ denote the latent regime assignment for layer $\ell$ of step $t$. For each layer, we compute the \emph{posterior responsibility} $\gamma_{t,\ell}(k) = P(z_{t,\ell}{=}k \mid \hat{h}_{t,\ell},\, c)$:
\begin{equation}
\gamma_{t,\ell}(k)
  = \frac{\pi^{(c)}_{\ell,k} \;\cdot\; b^{(c)}_k(\hat{h}_{t,\ell})}
         {\displaystyle\sum_{k'=1}^{K} \pi^{(c)}_{\ell,k'} \;\cdot\; b^{(c)}_{k'}(\hat{h}_{t,\ell})}.
\label{eq:gamma}
\end{equation}

\paragraph{M-step.}
For each category $c$ and regime $k$, the posteriors are accumulated over all steps and layers to update the GMM parameters:
\begin{align}
\mu^{(c)}_k
  &\leftarrow \sum_{t,\ell}\gamma_{t,\ell}(k)\,\hat{h}_{t,\ell}
    \;\Big/\; \sum_{t,\ell}\gamma_{t,\ell}(k),
  \label{eq:mstep_mu}\\
\sigma^{2(c)}_{k,d}
  &\leftarrow \frac{\sum_{t,\ell}\gamma_{t,\ell}(k)\,
    (\hat{h}_{t,\ell,d}-\mu^{(c)}_{k,d})^2}
    {\sum_{t,\ell}\gamma_{t,\ell}(k)},
  \label{eq:mstep_sigma}\\
\pi^{(c)}_k
  &\leftarrow \sum_{t,\ell \ge 2}\gamma_{t,\ell}(k)
    \;\Big/\; \sum_{k'}\sum_{t,\ell \ge 2}\gamma_{t,\ell}(k').
  \label{eq:mstep_pi}
\end{align}
The mean and variance updates use all layers, while the mixture weights $\pi^{(c)}$ are estimated from layers $\ell \ge 2$ only, because the first-layer prior is governed by the bridge matrix $J$.

The bridge matrix $J$ is updated by accumulating the joint posterior over consecutive steps:
\begin{equation}
J^{(c,c')}_{j,k}
  \leftarrow \frac{\sum_{t:\,c_{t-1}=c,\,c_t=c'}
    \gamma_{t-1,L}(j)\;\gamma_{t,1}(k)}
    {\sum_{k'}\sum_{t:\,c_{t-1}=c,\,c_t=c'}
    \gamma_{t-1,L}(j)\;\gamma_{t,1}(k')},
\label{eq:mstep_bridge}
\end{equation}
where $\gamma_{t-1,L}(j)$ is the exit posterior of the preceding step and $\gamma_{t,1}(k)$ is the entry posterior of the current step.

\subsection{Decoding}
The discrete regime assignment for each layer is decoded via per-layer maximum posteriori:
\begin{equation}
z^*_{t,\ell}
  = \operatorname*{arg\,max}_k \;\bigl[\log \pi^{(c)}_{\ell,k} + \log b^{(c)}_k(\hat{h}_{t,\ell})\bigr],
\label{eq:decode}
\end{equation}
where $\pi^{(c)}_{\ell,k}$ uses the bridge-modified weights at $\ell{=}1$ and the standard weights $\pi^{(c)}_k$ for $\ell \ge 2$.

\subsection{$K$ Selection}
The number of regimes $K$ is selected by sweeping a candidate set
$\{K_{\min},\dots,K_{\max}\}$: we run an EM process without bridge and compute the silhouette score~\cite{rousseeuw1987silhouettes}.
\begin{equation}
S(K) = \frac{1}{N}\sum_{t,\ell}
         \frac{\bar{d}_{\mathrm{near}}(t,\ell) - \bar{d}_{\mathrm{in}}(t,\ell)}{\max\{\bar{d}_{\mathrm{in}}(t,\ell),\bar{d}_{\mathrm{near}}(t,\ell)\}},
\end{equation}
where $\bar{d}_{\mathrm{in}}(t,\ell)$ is the mean intra-cluster distance
and $\bar{d}_{\mathrm{near}}(t,\ell)$ the mean nearest-cluster distance in PCA feature space.
The selected $K$ is then fully retrained with bridge.
\section{Use of AI Assistants.} 

AI Assistants were used solely as assistive tools for proofreading and improving clarity of writing.
\clearpage
\section{Results}
\label{app:results}

This section provides additional results. We first present the second-order transition results (Table~\ref{tab:second_order_no_un}). We then report the full first-order transition results, including all samples (Table~\ref{tab:first_order_all_2}), correct samples (Table~\ref{tab:first_order_correct}), incorrect samples (Table~\ref{tab:first_order_incorrect}), and their difference (correct $-$ incorrect) (Table~\ref{tab:first_order_diff}). We further summarize the overall statistics of reasoning trajectories (Table~\ref{tab:trajectory_summary}) and provide summary of failure mode behavior (Table~\ref{tab:overthinking-appendix}).

\begin{table}[h]
\centering
\resizebox{\columnwidth}{!}{
\begin{tabular}{lcccc}
\toprule
\textbf{Context} & \textbf{FA} & \textbf{SR} & \textbf{AC} & \textbf{UV} \\
\midrule
FA$\rightarrow$FA & 0.770$\pm$0.199 & 0.041$\pm$0.133 & 0.025$\pm$0.057 & 0.163$\pm$0.108 \\
FA$\rightarrow$SR & 0.244$\pm$0.117 & 0.515$\pm$0.210 & 0.180$\pm$0.211 & 0.060$\pm$0.056 \\
FA$\rightarrow$AC & 0.227$\pm$0.101 & 0.077$\pm$0.046 & 0.510$\pm$0.127 & 0.186$\pm$0.066 \\
FA$\rightarrow$UV & 0.506$\pm$0.234 & 0.067$\pm$0.036 & 0.178$\pm$0.101 & 0.249$\pm$0.142 \\
\midrule
SR$\rightarrow$FA & 0.558$\pm$0.212 & 0.100$\pm$0.052 & 0.171$\pm$0.141 & 0.171$\pm$0.114 \\
SR$\rightarrow$SR & 0.015$\pm$0.008 & 0.534$\pm$0.061 & 0.353$\pm$0.059 & 0.098$\pm$0.029 \\
SR$\rightarrow$AC & 0.033$\pm$0.013 & 0.175$\pm$0.040 & 0.660$\pm$0.063 & 0.131$\pm$0.039 \\
SR$\rightarrow$UV & 0.028$\pm$0.017 & 0.244$\pm$0.035 & 0.291$\pm$0.045 & 0.437$\pm$0.057 \\
\midrule
AC$\rightarrow$FA & 0.323$\pm$0.093 & 0.061$\pm$0.059 & 0.234$\pm$0.125 & 0.382$\pm$0.082 \\
AC$\rightarrow$SR & 0.046$\pm$0.031 & 0.236$\pm$0.033 & 0.627$\pm$0.073 & 0.090$\pm$0.030 \\
AC$\rightarrow$AC & 0.052$\pm$0.019 & 0.060$\pm$0.024 & 0.762$\pm$0.063 & 0.126$\pm$0.044 \\
AC$\rightarrow$UV & 0.056$\pm$0.022 & 0.102$\pm$0.019 & 0.390$\pm$0.056 & 0.451$\pm$0.051 \\
\midrule
UV$\rightarrow$FA & 0.418$\pm$0.130 & 0.041$\pm$0.047 & 0.060$\pm$0.068 & 0.481$\pm$0.125 \\
UV$\rightarrow$SR & 0.044$\pm$0.027 & 0.333$\pm$0.034 & 0.447$\pm$0.052 & 0.175$\pm$0.042 \\
UV$\rightarrow$AC & 0.054$\pm$0.021 & 0.091$\pm$0.016 & 0.583$\pm$0.068 & 0.271$\pm$0.063 \\
UV$\rightarrow$UV & 0.053$\pm$0.021 & 0.125$\pm$0.029 & 0.286$\pm$0.044 & 0.536$\pm$0.065 \\
\bottomrule
\end{tabular}
}
\caption{Second-order transition probabilities over semantic categories. Values are reported as mean $\pm$ standard deviation across model/dataset.}
\label{tab:second_order_no_un}
\end{table}

\begin{table}[h]
\centering
\resizebox{\columnwidth}{!}{
\begin{tabular}{lcccc}
\toprule
\textbf{From} $\backslash$ \textbf{To} & \textbf{FA} & \textbf{SR} & \textbf{AC} & \textbf{UV} \\
\midrule
FA & $0.466\pm0.170$ & $0.076\pm0.065$ & $0.131\pm0.097$ & $0.326\pm0.098$ \\
SR & $0.038\pm0.022$ & $0.419\pm0.053$ & $0.445\pm0.066$ & $0.098\pm0.029$ \\
AC & $0.060\pm0.019$ & $0.079\pm0.029$ & $0.715\pm0.073$ & $0.147\pm0.053$ \\
UV & $0.124\pm0.111$ & $0.119\pm0.024$ & $0.320\pm0.061$ & $0.438\pm0.077$ \\
\bottomrule
\end{tabular}
}
\caption{First-order transition matrix over semantic categories (All trajectories). 
Values are reported as mean $\pm$ standard deviation across model/dataset.}
\label{tab:first_order_all_2}
\end{table}

\begin{table}[h]
\centering
\resizebox{\columnwidth}{!}{
\begin{tabular}{lcccc}
\toprule
\textbf{From} $\backslash$ \textbf{To} & \textbf{FA} & \textbf{SR} & \textbf{AC} & \textbf{UV} \\
\midrule
FA & $0.454\pm0.140$ & $0.085\pm0.066$ & $0.131\pm0.076$ & $0.331\pm0.087$ \\
SR & $0.034\pm0.021$ & $0.416\pm0.062$ & $0.464\pm0.076$ & $0.086\pm0.030$ \\
AC & $0.061\pm0.015$ & $0.073\pm0.028$ & $0.734\pm0.063$ & $0.132\pm0.045$ \\
UV & $0.120\pm0.100$ & $0.117\pm0.024$ & $0.342\pm0.060$ & $0.421\pm0.067$ \\
\bottomrule
\end{tabular}
}
\caption{First-order transition matrix over semantic categories for correct trajectories. 
Values are reported as mean $\pm$ standard deviation across model/dataset.}
\label{tab:first_order_correct}
\end{table}

\begin{table}[h]
\centering
\resizebox{\columnwidth}{!}{
\begin{tabular}{lcccc}
\toprule
\textbf{From} $\backslash$ \textbf{To} & \textbf{FA} & \textbf{SR} & \textbf{AC} & \textbf{UV} \\
\midrule
FA & $0.487\pm0.206$ & $0.062\pm0.063$ & $0.131\pm0.117$ & $0.319\pm0.118$ \\
SR & $0.043\pm0.027$ & $0.424\pm0.056$ & $0.413\pm0.069$ & $0.119\pm0.031$ \\
AC & $0.057\pm0.024$ & $0.088\pm0.029$ & $0.683\pm0.083$ & $0.172\pm0.059$ \\
UV & $0.129\pm0.120$ & $0.122\pm0.027$ & $0.283\pm0.060$ & $0.466\pm0.084$ \\
\bottomrule
\end{tabular}
}
\caption{First-order transition matrix over semantic categories for incorrect trajectories. 
Values are reported as mean $\pm$ standard deviation across seeds.}
\label{tab:first_order_incorrect}
\end{table}

\begin{table}[h]
\centering
\resizebox{\columnwidth}{!}{
\begin{tabular}{lcccc}
\toprule
\textbf{From} $\backslash$ \textbf{To} & \textbf{FA} & \textbf{SR} & \textbf{AC} & \textbf{UV} \\
\midrule
FA & $-0.034\pm0.137$ & $+0.022\pm0.031$ & $-0.000\pm0.085$ & $+0.012\pm0.072$ \\
SR & $-0.009\pm0.022$ & $-0.009\pm0.061$ & $+0.051\pm0.056$ & $-0.033\pm0.027$ \\
AC & $+0.004\pm0.015$ & $-0.015\pm0.014$ & $+0.051\pm0.028$ & $-0.041\pm0.021$ \\
UV & $-0.009\pm0.040$ & $-0.005\pm0.013$ & $+0.059\pm0.038$ & $-0.045\pm0.047$ \\
\bottomrule
\end{tabular}
}
\caption{Difference in first-order transition probabilities between correct and incorrect trajectories (Correct $-$ Incorrect). Values are reported as mean $\pm$ standard deviation across model/dataset.}
\label{tab:first_order_diff}
\end{table}

\begin{table}[h]
\centering
\resizebox{\columnwidth}{!}{
\begin{tabular}{lcc}
\toprule
\textbf{Metric} & \textbf{Category} & \textbf{Value} \\
\midrule
\multirow{2}{*}{Avg Path Length}
 & Correct   & $149.1$ \\
 & Incorrect & $265.1$ \\
\midrule
\multirow{4}{*}{Stationary Distribution ($\pi$)}
 & FA & $0.125\pm0.114$ \\
 & SR & $0.132\pm0.044$ \\
 & AC & $0.507\pm0.113$ \\
 & UV & $0.235\pm0.061$ \\
\midrule
\multirow{3}{*}{Expected Steps to FA}
 & AC & $22.4\pm12.6$ \\
 & SR & $22.9\pm12.7$ \\
 & UV & $21.7\pm13.1$ \\
\bottomrule
\end{tabular}
}
\caption{Summary statistics of reasoning trajectories. Values are reported as mean $\pm$ standard deviation across runs.}
\label{tab:trajectory_summary}
\end{table}

\begin{table*}[t]
\centering
\resizebox{\textwidth}{!}{
\begin{tabular}{ll*{4}{c}*{4}{c}*{4}{c}*{3}{c}|c}
\toprule
& & \multicolumn{4}{c}{Stratos} & \multicolumn{4}{c}{OpenThinker} & \multicolumn{4}{c}{Qwen} & \multicolumn{3}{c|}{Nemotron} & \\
\cmidrule(lr){3-6} \cmidrule(lr){7-10} \cmidrule(lr){11-14} \cmidrule(lr){15-17}
& & MATH500 & GPQA-D & AIME24 & WebInstruct
& MATH500 & GPQA-D & AIME24 & WebInstruct
& MATH500 & GPQA-D & AIME24 & WebInstruct
& MATH500 & GPQA-D & WebInstruct & All \\
\midrule
\multicolumn{18}{l}{\textit{Regime $L_2$ divergence (long fail vs.\ short fail, pooled $\pm$ std)}} \\[2pt]
& AC & \ms{0.44}{.15} & \ms{1.34}{.06} & \ms{1.10}{.13} & \ms{0.65}{.13} & \ms{0.61}{.17} & \ms{1.45}{.39} & \ms{0.89}{.22} & \ms{0.75}{.04} & \ms{0.92}{.09} & \ms{0.73}{.08} & \ms{2.16}{.37} & \ms{0.49}{.05} & \ms{1.09}{.13} & \ms{\textbf{1.71}}{.09} & \ms{0.73}{.08} & 1.00 \\
& UV & \ms{1.41}{.20} & \ms{1.10}{.31} & \ms{1.51}{.53} & \ms{1.55}{.59} & \ms{1.35}{.08} & \ms{2.34}{.36} & \ms{1.63}{.12} & \ms{1.14}{.24} & \ms{1.13}{.13} & \ms{0.53}{.02} & \ms{1.03}{.52} & \ms{1.01}{.15} & \ms{1.92}{.21} & \ms{0.55}{.10} & \ms{0.90}{.09} & 1.27 \\
& FA & \ms{\textbf{2.36}}{.29} & \ms{0.91}{.30} & \ms{1.31}{.04} & \ms{\textbf{2.05}}{.62} & \ms{\textbf{4.63}}{.47} & \ms{\textbf{2.53}}{.19} & \ms{\textbf{3.30}}{0} & \ms{\textbf{5.25}}{.29} & \ms{\textbf{1.52}}{.09} & \ms{0.65}{.12} & \ms{\textbf{2.40}}{.17} & \ms{1.04}{.12} & \ms{\textbf{3.15}}{.81} & \ms{1.47}{.59} & \ms{1.46}{.96} & \textbf{2.27} \\
& SR & \ms{1.19}{.14} & \ms{\textbf{2.06}}{1.50} & \ms{\textbf{1.56}}{.78} & \ms{1.30}{.04} & \ms{2.31}{.14} & \ms{0.74}{.07} & \ms{1.14}{1.88} & \ms{1.74}{.24} & \ms{1.37}{.07} & \ms{\textbf{0.93}}{.14} & \ms{2.21}{1.21} & \ms{\textbf{1.14}}{.15} & \ms{1.99}{.38} & \ms{1.38}{.21} & \ms{\textbf{1.75}}{.13} & 1.52 \\
\midrule
\multicolumn{18}{l}{\textit{FA visit metrics (median [Q1, Q3] per configuration; All column from pooled data)}} \\[2pt]
\multirow{3}{*}{\shortstack[l]{FA\\pos}}
& long & 0.29 & 0.25 & 0.36 & 0.21 & 0.28 & 0.30 & 0.33 & 0.24 & 0.37 & 0.33 & 0.61 & 0.23 & 0.29 & 0.46 & 0.27 & 0.28\,[.14,\,.55] \\
& short & 0.72 & 0.45 & 0.44 & 0.59 & 0.59 & 0.59 & 0.74 & 0.53 & 0.37 & 0.55 & 0.50 & 0.57 & 0.73 & 0.69 & 0.73 & 0.58\,[.38,\,.80] \\
\midrule
\multirow{3}{*}{\shortstack[l]{FA\\ep}}
& long & 7 & 7 & 5 & 10 & 8 & 7 & 7 & 10 & 8 & 7 & 8 & 13 & 10 & 5 & 13 & 8\,[5,\,15] \\
& short & 2 & 3 & 7 & 3 & 3 & 3 & 4 & 3 & 4 & 4 & 4 & 3 & 2 & 2 & 2 & 3\,[2,\,5] \\
\midrule
\multirow{3}{*}{\shortstack[l]{Waste\\frac}}
& long & 0.95 & 0.89 & 0.97 & 0.90 & 0.93 & 0.85 & 0.96 & 0.88 & 0.91 & 0.86 & 0.89 & 0.87 & 0.92 & 0.90 & 0.88 & 0.90\,[.82,\,.95] \\
& short & 0.67 & 0.73 & 0.80 & 0.67 & 0.70 & 0.71 & 0.62 & 0.71 & 0.81 & 0.77 & 0.81 & 0.62 & 0.60 & 0.75 & 0.62 & 0.71\,[.33,\,.81] \\
\bottomrule
\end{tabular}
}
\caption{Overthinking analysis. FA visit metrics show median [Q1, Q3]. $L_2$ values show pooled $\pm$ cross-seed std.}
\label{tab:overthinking-appendix}
\end{table*}

\begin{figure}[t]
    \centering
    \includegraphics[width=\linewidth]{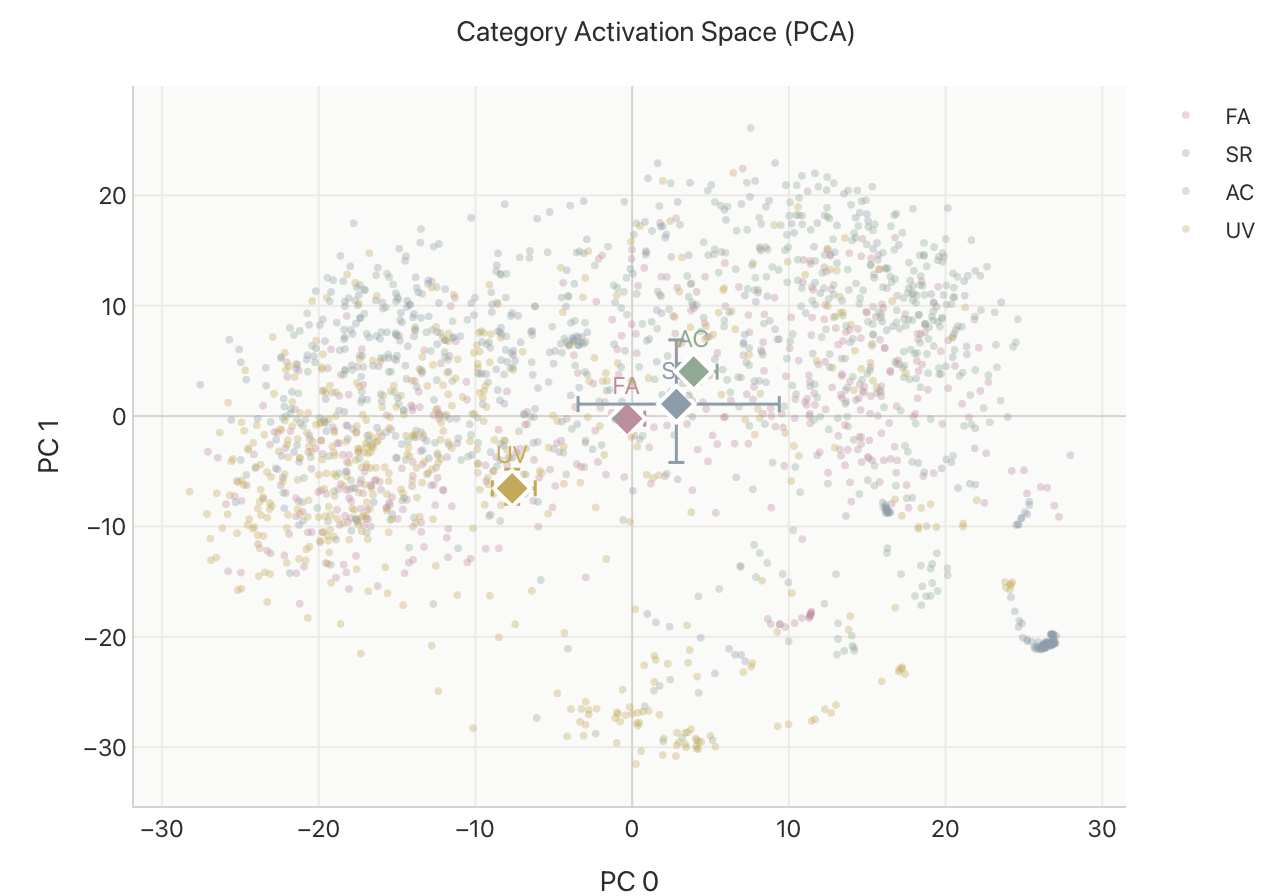}
    \caption{2D PCA projection of per-step activations colored by category. Diamond markers = category centroids. Separation between clusters indicates distinct hidden-state representations per category.}
    \label{fig:cas_app}
\end{figure}

\begin{figure*}
    \centering
    \includegraphics[width=\linewidth]{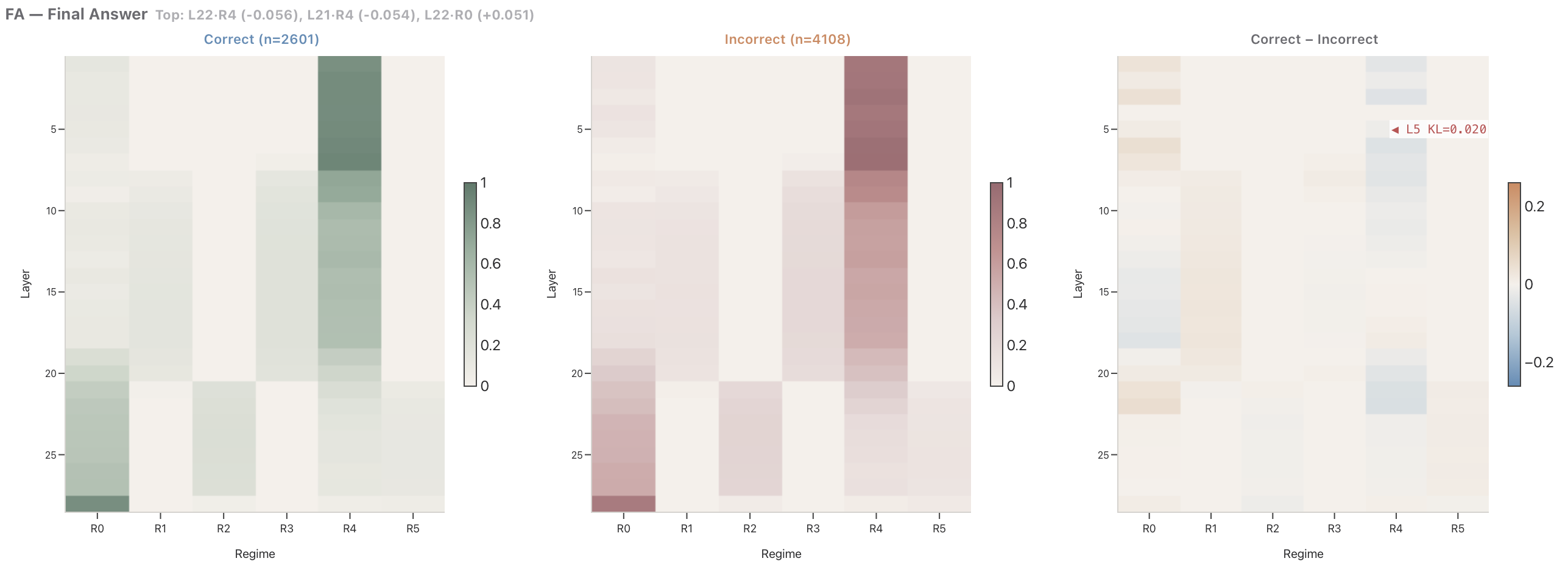}
    \caption{Implicit Stage Soft Profile: Final Answer}
    \label{fig:sp-fa}
\end{figure*}

\begin{figure*}
    \centering
    \includegraphics[width=\linewidth]{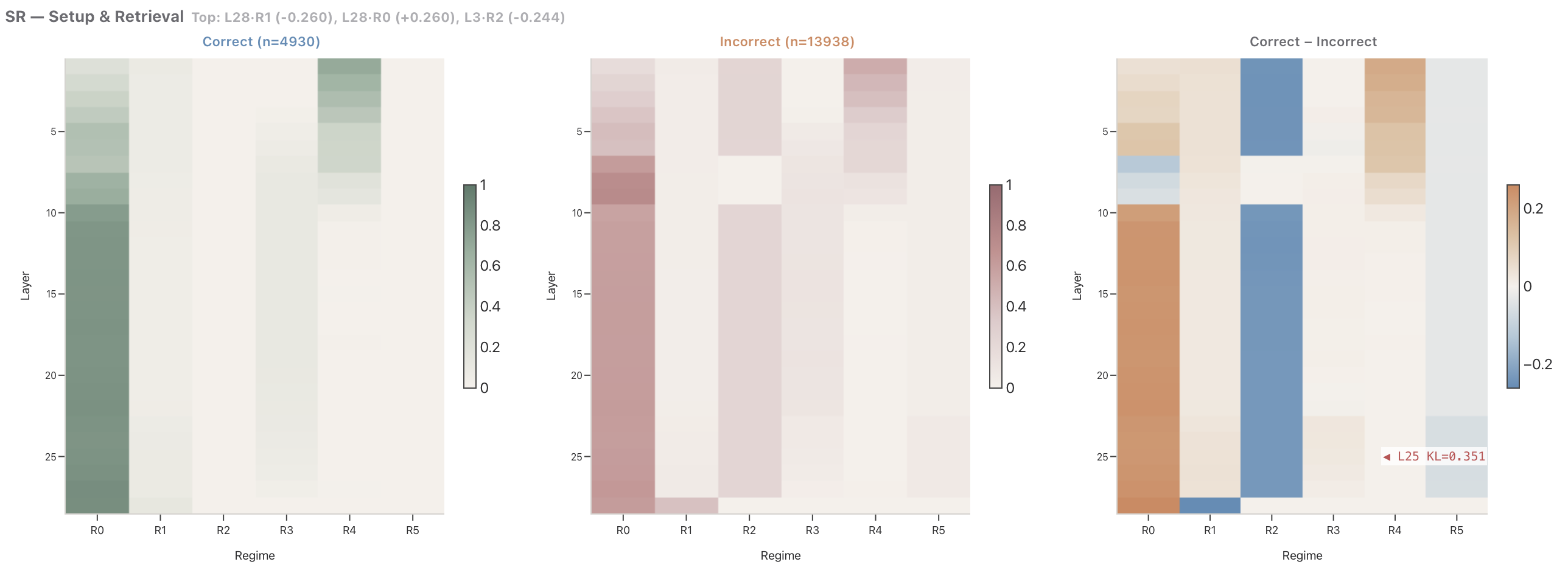}
    \caption{Implicit Stage Soft Profile: Setup \& Retrieval}
    \label{fig:sp-sr}
\end{figure*}

\begin{figure*}
    \centering
    \includegraphics[width=\linewidth]{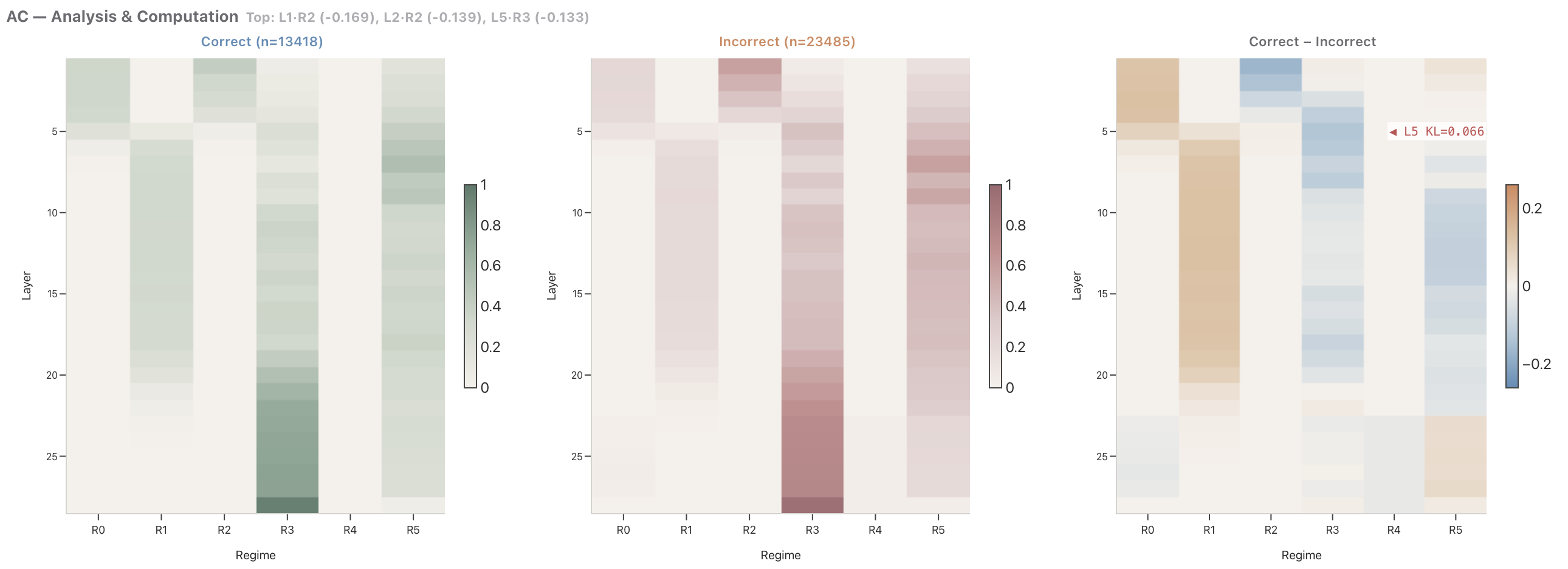}
    \caption{Implicit Stage Soft Profile: Analysis \& Computation}
    \label{fig:sp-ac}
\end{figure*}

\begin{figure*}
    \centering
    \includegraphics[width=\linewidth]{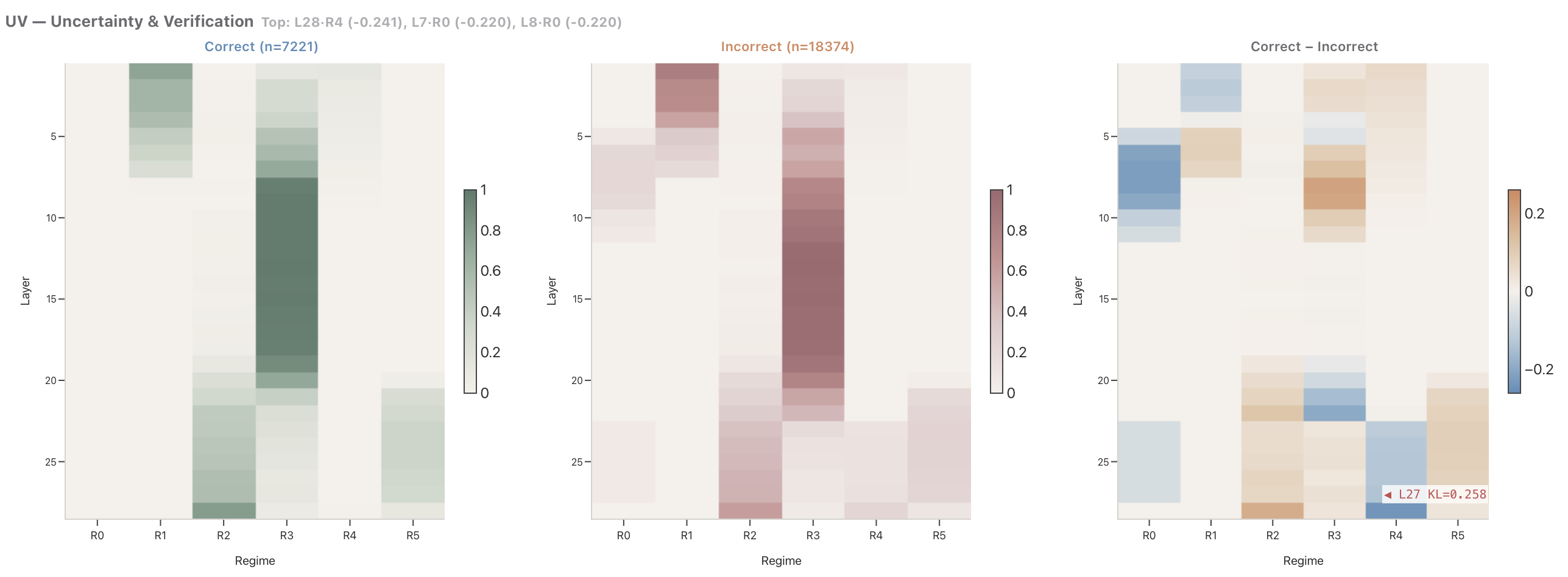}
    \caption{Implicit Stage Soft Profile: Uncertainty \& Verification}
    \label{fig:sp-uv}
\end{figure*}

We also provide Bespoke-Stratos-7B on GPQA-Diamond benchmark as illustrative example, including category activation space (Figure~\ref{fig:cas_app}), Posterior probability difference (Figure~\ref{fig:sp-fa} to \ref{fig:sp-uv}) and Explicit Bridge matrices (Table~\ref{tab:app-exit_regime_transitions}).

\begin{table}[h]
\centering
\resizebox{\columnwidth}{!}{%
\begin{tabular}{c|cccc|cccc|cccc|cccc}
\toprule
& \multicolumn{4}{c|}{\textbf{Source: FA}}
& \multicolumn{4}{c|}{\textbf{Source: SR}}
& \multicolumn{4}{c|}{\textbf{Source: AC}}
& \multicolumn{4}{c}{\textbf{Source: UV}} \\
\cmidrule(lr){2-5} \cmidrule(lr){6-9} \cmidrule(lr){10-13} \cmidrule(lr){14-17}
\textbf{Regime}
& FA & SR & AC & UV
& FA & SR & AC & UV
& FA & SR & AC & UV
& FA & SR & AC & UV \\
\midrule
R0 & 43 & 8 & 18 & 32 & 5 & 35 & 46 & 14 & 0 & 0 & 87 & 13 & 0 & 0 & 0 & 100 \\
R1 & 0 & 0 & 0 & 0 & 0 & 80 & 9 & 11 & 0 & 0 & 99 & 1 & 12 & 49 & 17 & 23 \\
R2 & 71 & 6 & 13 & 9 & 0 & 0 & 0 & 0 & 0 & 0 & 0 & 100 & 7 & 15 & 28 & 50 \\
R3 & 0 & 0 & 0 & 0 & 4 & 69 & 25 & 3 & 5 & 12 & 61 & 22 & 4 & 14 & 36 & 47 \\
R4 & 38 & 7 & 21 & 34 & 6 & 22 & 62 & 9 & 23 & 13 & 46 & 18 & 0 & 14 & 25 & 62 \\
R5 & 92 & 0 & 0 & 8 & 0 & 0 & 0 & 0 & 4 & 8 & 68 & 19 & 3 & 20 & 28 & 49 \\
\bottomrule
\end{tabular}
}
\caption{Explicit Bridge: P(target category | source category, exit regime).}
\label{tab:app-exit_regime_transitions}
\end{table}

\begin{figure*}
    \centering
    \includegraphics[width=\linewidth]{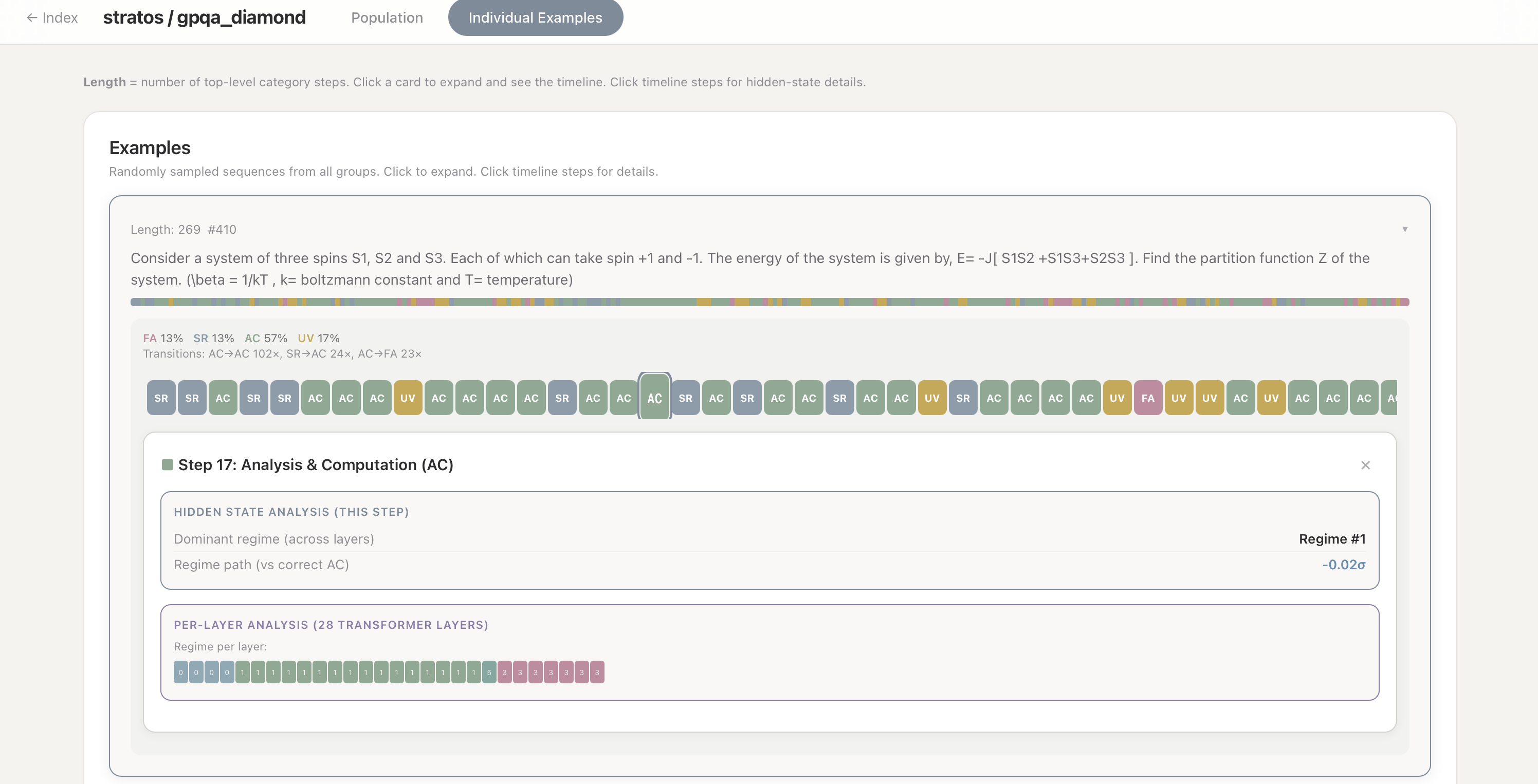}
    \caption{Individual Examples.}
    \label{fig:placeholder}
\end{figure*}

\end{document}